%% file: arxiv.tex
\documentclass[11pt, a4paper, logo, copyright]{deepmind}
\usepackage[round]{natbib}

\usepackage{microtype}
\usepackage{graphicx}
\usepackage{subfigure}
\usepackage{booktabs} 
\usepackage{hyperref}

\usepackage{amsmath}
\usepackage{amssymb}
\usepackage{mathtools}
\usepackage{amsthm}
\usepackage{algorithm}
\usepackage{algorithmic}

\usepackage[capitalize,noabbrev]{cleveref}

\DeclareMathOperator*{\MP}{MP}
\DeclareMathOperator*{\VI}{VI}
\DeclareMathOperator*{\new}{new}
\DeclareMathOperator*{\old}{old}
\DeclareMathOperator*{\argmin}{argmin}
\DeclareMathOperator*{\argmax}{argmax}
\DeclareMathOperator*{\MAP}{MAP}

\DeclareMathOperator*{\Alg1}{Alg1}

\usepackage{tikz}
\usetikzlibrary{bayesnet}

\title{Learning noisy-OR Bayesian Networks with Max-Product Belief Propagation}
\author[1]{Antoine Dedieu}
\author[1]{Guangyao Zhou}
\author[1]{Dileep George}
\author[1]{Miguel L\'{a}zaro-Gredilla}
\affil[1]{DeepMind}
\correspondingauthor{adedieu@deepmind.com}

\begin{abstract}
Noisy-OR Bayesian Networks (BNs) are a family of probabilistic graphical models which express rich statistical dependencies in binary data. 
Variational inference (VI) has been the main method proposed to learn noisy-OR BNs with complex latent structures \citep{jaakkola1999variational,ji2020variational,buhai2020empirical}.
However, the proposed VI approaches either (a) use a recognition network with standard amortized inference that cannot induce ``explaining-away''; or (b) assume a simple mean-field (MF) posterior which is vulnerable to bad local optima. Existing MF VI methods also update the MF parameters sequentially which makes them inherently slow. 
In this paper, we propose parallel max-product as an alternative algorithm for learning noisy-OR BNs with complex latent structures and we derive a fast stochastic training scheme that scales to large datasets.
We evaluate both approaches on several benchmarks where VI is the state-of-the-art and show that our method 
(a) achieves better test performance than \citet{ji2020variational} for learning noisy-OR BNs with hierarchical latent structures on large sparse real datasets;
(b) recovers a higher number of ground truth parameters than \citet{buhai2020empirical} from cluttered synthetic scenes; and
(c) solves the 2D blind deconvolution problem from \citet{lazaro2021perturb} and variants---including binary matrix factorization---while VI catastrophically fails and is up to two orders of magnitude slower.
\end{abstract}

\begin{document}
\maketitle
\input{main_text_arxiv}
\end{document}

%% file: main_text_arxiv.tex
\section{Introduction}
Probabilistic graphical models (PGMs) propose a rigorous and elegant way to represent the full joint probability density function of high-dimensional data and to express assumptions about its hidden structure. Learning and inference algorithms let us analyze data under those assumptions and recover the hidden structure that best explains our observations. However, performing exact inference in complex PGMs is often intractable. To mitigate this problem, several techniques have been proposed for approximate inference, among which a popular one is variational inference (VI) \citep{wainwright2008graphical, bishop2006pattern}. 

In this paper, we consider directed acyclic PGMs---also named Bayesian networks (BNs)---with binary variables and noisy-OR conditional distribution \citep{pearl1988probabilistic}. The resulting noisy-OR BNs have been used for medical diagnosis \citep{jaakkola1999variational}, data compression \citep{vsingliar2006noisy}, text mining \citep{liu2016representing}, and more recently overparametrized learning \citep{buhai2020empirical} and topic modeling on large sparse datasets \citep{ji2020variational}. Noisy-OR BNs have an intractable posterior: most of the aforementioned applications rely on VI for approximate inference. Some existing VI methods \citep{buhai2020empirical} use a recognition network and amortize the approximate inference via a single forward pass, which cannot induce ``explaining-away'' (see Section \ref{sec:noisy-or}). In contrast, \citet{jaakkola1999variational,vsingliar2006noisy,ji2020variational} assume a mean-field (MF) posterior, which is vulnerable to bad local optima. These existing MF methods also update the MF parameters sequentially---i.e. one by one---which is prohibitively slow. To scale MF VI, \citet{ji2020variational} propose a local heuristic that updates fewer MF parameters (see Section \ref{sec:related}). However, their approach only applies to large sparse datasets (i.e. most of the observations are $0$s).

In this work, we propose a fast and efficient stochastic scheme for learning noisy-OR BNs that use the parallel max-product (MP) algorithm \citep{pearl1988probabilistic,murphy2013loopy} as an alternative to VI.
Similar to \citet{ji2020variational}, our method supports multi-layered noisy-OR networks and relies on stochastic optimization \citep{robbins1951stochastic} for scaling. However, (a) contrary to  \citet{ji2020variational}, our approach runs in parallel which allows it to scale to large dense datasets; and (b) in contrast with \citet{buhai2020empirical}, our method induces explaining-away. We show that our approach efficiently explores the parameters space, which allows better performance on the experiments of \citet{ji2020variational}. We additionally show that several challenging problems including (a) binary matrix factorization; (b) the noisy-OR BNs experiments from \citet{buhai2020empirical}; and (c) the complex 2D blind deconvolution problem from \citet{lazaro2021perturb} can be expressed as learning problems in noisy-OR BNs, for which MP outperforms VI while being up to two orders of magnitude faster. Our code is written in \texttt{JAX} \citep{jax2018github} and will be made public after publication.

The rest of this paper is organized as follows. Section \ref{sec:noisy-or} reviews noisy-OR BNs, while Section \ref{sec:related} discusses existing learning methods for these models. Section \ref{sec:MP-recent-work} introduces the max-product algorithm, which Section \ref{sec:training} integrates into our training scheme for BNs. Finally, Section \ref{sec:exp} compares our method with VI in a wide variety of experiments.

\begin{figure*}[b!]
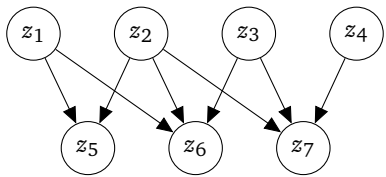

\centering
\tikz{ 
    \node[latent] (x1) {$z_5$}; 
    \node[latent, right=.7 of x1] (x2) {$z_6$}; 
    \node[latent, right=.7 of x2] (x3) {$z_7$}; 
    \node[latent, above=0.8 of x1, xshift=7mm] (h2) {$z_2$}; 
    \node[latent, above=0.8 of x1, left=.7 of h2] (h1) {$z_1$}; 
    \node[latent, above=0.8 of x1, right=.7 of h2] (h3) {$z_3$}; 
    \node[latent, above=0.8 of x1, right=.7 of h3] (h4) {$z_4$}; 
    \path[->,draw]
    (h1) edge node {} (x1)
    (h1) edge node {} (x2)
    (h2) edge node {} (x1)
    (h2) edge node {} (x2)
    (h2) edge node {} (x3)
    (h3) edge node {} (x2)
    (h3) edge node {} (x3)
    (h4) edge node {} (x3)
    
}
\caption{Two-layers noisy-OR Bayesian network with four hidden and three visible nodes. The leak node ($z_0=1$) is not shown.}
\label{fig:BN}
\end{figure*}

\section{Noisy-OR Bayesian networks} \label{sec:noisy-or}
Given binary observations $x \in \{0,1\}^p$, we model its statistical dependencies using binary BNs \citep{koller2009probabilistic} as in Figure \ref{fig:BN}. 
The nodes in the graph are divided into $p$ visible nodes---which are the leaves---and $m$ hidden nodes. Each visible (resp. hidden) node $i$ is associated with a binary random variable $x_i$ (resp. $h_i$). We denote $h=(h_1, \ldots, h_m)$ and $x=(x_1, \ldots, x_p)$. Similar to \citet{ji2020variational}, we introduce a leak node $0$ that connects to all the nodes, whose variable $z_0$ is always active, i.e. $z_0=1$. The leak node allows any active variable to be explained by other factors than its parents. For convenience, we denote $z=(z_0, h, x)$ the vector of all variables: $h=(z_1, \ldots, z_m)$ and $x=(z_{m+1}, \ldots z_{m+p})$.

Let $\mathcal{P}(i)$ be the set of parents (excluding the leak node) of the node $i\ge1$. The activation probability of the variable $z_i$ is given by the noisy-OR conditional distribution
\begin{align}\label{noisyOR}
&p(z_i=0 ~|~ z_{\mathcal{P}(i)}, ~\Theta) 
=\exp \biggl(-\theta_{0 \to i} - \sum_{k \in \mathcal{P}(i)} \theta_{k \to i} z_k \biggr)
\end{align}
where $\theta_{0 \to i} \ge 0, ~\theta_{k \to i} \ge 0, ~\forall k \in \mathcal{P}(i)$ and $\Theta$ is the vector collecting all these parameters. This conditional distribution possesses three important properties. First, if all the parents are inactive, the activation probability is given by the leak node: $p(z_i=0 ~|~ z_{\mathcal{P}(i)}=0, ~\Theta)= \exp (- \theta_{0 \to i} )$. As in \citet{buhai2020empirical}, we refer to $1- \exp (- \theta_{0 \to i})$ as the ``prior probability'' when $\mathcal{P}(i)$ is empty and the ``noise probability'' otherwise. Second, if only the variable $k$ is connected to the variable $i$ and there is no leak node, $p(z_i=0 ~|~ z_k=1, ~\Theta)=\exp (- \theta_{k \to i} )$---which we refer to as the ``failure probability''. Finally, noisy-OR BNs can induce ``explaining-away'': explaining-away creates competition between a-priori unlikely causes, which allows inference to pick the smallest subset of causes that explain the effects.



\section{Related Work} \label{sec:related}
The QMR-DT network \citep{jaakkola1999variational} is one of the first models which exploits the properties of noisy-OR BNs. It consists of a two-layer bipartite graph created by domain experts which models how $600$ diseases explain $4,000$ findings. The probability of a finding given diseases is expressed by Equation \eqref{noisyOR}. After learning, the QMR-DT  network is used to infer the probabilities of different diseases given a set of observed symptoms. For approximate inference in the intractable noisy-OR BN, the authors assumed a MF posterior---which can induce explaining-away (see Section \ref{sec:noisy-or})---and introduced a family of variational bounds as well as a heuristic to increase the graph sparsity.

Other approaches have been proposed for learning bipartite noisy-OR BNs. \citet{vsingliar2006noisy} introduced a variational EM procedure that exploits the bounds of \citet{jaakkola1999variational} while assuming a fully connected graph. \citet{halpern2013unsupervised} proposed a method of moments that requires the graph to be sparse. \citet{liu2016representing} introduced a Monte-Carlo EM algorithm that requires a large number of sampling steps for good performance. None of these methods would scale to large datasets.

Recently, \citet{buhai2020empirical} discussed the effect of overparameterization in PGMs and showed that, on synthetic datasets, increasing the number of latent variables of noisy-OR BNs improves their performance at recovering the ground truth parameters. Their method considered VI with a recognition network. However, the authors amortize the inference via a single forward pass: inference results in picking all causes that are consistent with the effects and cannot induce explaining-away (see Section \ref{sec:noisy-or}).

In another recent work, \citet{ji2020variational} proposed a stochastic variational training algorithm for noisy-OR BNs. The authors assumed a MF posterior and extended the bounds of \citet{jaakkola1999variational}. For scalability, the authors introduced ``local models'': they only update the variational parameters associated with the ancestors of the active visible variables. They showed that this is equivalent to optimizing a constrained variational bound, and derived state-of-the-art performance for multi-layered BNs on large sparse real datasets, while significantly outperforming \citet{liu2016representing}.

The method we propose in Section \ref{sec:training} for learning noisy-OR BNs has the same appealing properties as \citet{ji2020variational}: it induces explaining-away, it supports multi-layered graphs and it scales to large sparse datasets. In addition, (a) it is faster as it runs parallel max-product; (b) it also scales to large dense datasets; and (c) it considers a richer posterior than MF VI which allows it to find better local optima. 

\section{Background on max-product}\label{sec:MP-recent-work}

We first review the parallel max-product algorithm. We then discuss how this algorithm can be used for sampling in PGMs, and how it can be easily accelerated on GPUs.

\subsection{Max-product message passing}\label{sec:max-product}

We consider a PGM with variables $z$ described by a set of $A$ factors $\{\psi_a^\top\phi_a(z^a)\}_{a=1}^A$ and $I$ unary terms $\{\lambda_i^\top\eta_i(z_i)\}_{i=1}^I$. $z^a$ is the vector of variables used in the factor $a$, $\psi_a$ is a vector of factor parameters and $\phi_a(z^a)$ is a vector of factor sufficient statistics. For the unary terms, the sufficient statistics are the indicator functions $\eta_i(x_i) = (\mathbf{1}(x_i=0), \mathbf{1}(x_i=1))$. For a Bayesian network, a factor involving $z^a=(z_a, z_{\mathcal{P}(a)}, z_0)$ is defined for the $a$th variable. The corresponding $\psi_a$ can be derived from the parameters $\{\theta_{0 \to a}\} \cup \big\{ \theta_{k \to a} \big\}_{k \in \mathcal{P}(a)}$ defined in Equation \eqref{noisyOR}.

The energy of the model can be expressed as $E(z) = -\sum_{a=1}^A \psi_a^\top\phi_a(z^a) -\sum_{i=1}^I \lambda_i^\top\eta_i(z_i)$ or, collecting the parameters and sufficient statistics in corresponding vectors, $E(z) =- \Psi^\top \Phi(z)  - \Lambda^\top \eta(z) $. The probability of a configuration $z$ satisfies $p(z) \propto \exp(- E(z))$. The maximum a posteriori (MAP) problem consists in finding the variable assignment with the lowest energy, that is 
\begin{equation}\label{MAP}
z^{\MAP} \in \argmin_z E(z)=\argmax_z~ \Psi^\top \Phi(z) + \Lambda^\top \eta(z) 
\end{equation}
The max-product algorithm estimates this solution by iterating the fixed-point updates for $N_{\MP}$ iterations:
\begin{align}\label{eq:bpupdates}
m_{i \rightarrow a}(z_i) &= \lambda_i^\top\eta_i(z_i) + \sum_{b\in \text{nb}(i)\backslash a} m_{b \rightarrow i}(z_i) \\
m_{a \rightarrow i}(z_i) &= \max_{z_{k\backslash i}} \bigg\{ \psi_a^\top\phi_a(z^a) + \sum_{k\in \text{nb}(a)\backslash i} m_{k \rightarrow a}(z_k) \bigg\}\nonumber
\end{align}
where $\text{nb}(\cdot)$ denotes the neighbors of a factor or variable. Equations \eqref{eq:bpupdates} are derived by setting the gradients of the Lagrangian of the Bethe free energy to $0$---see \citet{wainwright2008graphical}. $m_{i \rightarrow a}(z_i)$ (resp. $m_{a \rightarrow i}(z_i)$) are called the ``messages'' from variables to factors (resp. from factors to variables): max-product is a ``message-passing'' algorithm. After $N_{\MP}$ iterations of Equation \eqref{eq:bpupdates}, max-product estimates the solution to Problem \eqref{MAP} by
$$
z_i = \argmax_c \bigg\{ \lambda_i^\top\eta_i(z_i=c) + \sum_{b\in \text{nb}(i)} m_{b \rightarrow i}(z_i=c) \bigg\}, ~ \forall i.
$$
MP is guaranteed to converge in trees like BNs \citep{weiss1997belief}. A damping factor $\alpha \in (0, 1)$ in the updates can be used to improve convergence, so that
$m^{\new}_{a \rightarrow i}(z_i) = \alpha m_{a \rightarrow i}(z_i) + (1-\alpha) m^{\old}_{a \rightarrow i}(z_i)$. $\alpha=0.5$ offers a good trade-off between accuracy and speed in most cases.

\textbf{Max-product in BNs: } The noisy-OR factor in Equation \eqref{noisyOR} connects the variables $\{z_i\} \cup\{z_0\} \cup z_{\mathcal{P}(i)}$ and has $2^{2 + | \mathcal{P}(i)|}$ valid configurations. At first sight, the max-product updates in Equations \eqref{eq:bpupdates} have an exponential complexity in $\mathcal{O}(2^{| \mathcal{P}(i)|})$. To scale to large factors, we derive in Appendix \ref{appendix:pgmax-noisy-or} an equivalent representation of this noisy-OR factor for which the updates have a linear complexity  $\mathcal{O}(| \mathcal{P}(i)|)$.

\subsection{Sampling in PGMs via perturb-and-max-product}\label{sec:pmp}

In this work, we are interested in answering two types of inference queries in PGMs: MAP queries as in Problem \eqref{MAP} and sampling queries. The perturb-and-MAP framework \citep{papandreou2011perturb} unifies these two types of queries by considering the problem:
\begin{equation}\label{pmap}
\argmax_z \left\{ \Psi^\top\Phi(z)+(\Lambda + T ~ \varepsilon)^\top\eta(z) \right\}
\end{equation}
where $\varepsilon \in \mathbb{R}^{2I}$  is a perturbation vector added to the vector of unaries $\Lambda$, and $T$ is a temperature parameter. When $T=0$, Problem \eqref{pmap} is the MAP Problem \eqref{MAP}. When $T=1$, \citet{papandreou2011perturb} showed that if the entries of $\varepsilon$ are independently drawn from a Gumbel distribution, the solution of Problem \eqref{pmap} approximates a sample from the PGM distribution. \citet{lazaro2021perturb} recently showed state-of-the-art learning and sampling performance on several PGMs including Ising models and Restricted Boltzmann Machines by using max-product to solve Problem \eqref{pmap}. We use their method, named perturb-and-max-product (PMP), in the rest of this paper.

\subsection{Accelerating max-product on GPUs}
Recently \citet{zhou2022pgmax} open-sourced \texttt{PGMax}, a \texttt{Python} package to run GPU-accelerated parallel max-product on general factor graphs with discrete variables. The authors showed timing improvements of two to three orders of magnitude compared with alternatives. We use this package to solve the families of perturbed MAP Problems \eqref{pmap} for noisy-OR BNs, while performing GPU-accelerated message updates with linear complexity (see Appendix \ref{appendix:pgmax-noisy-or}).

\section{Noisy-OR Bayesian Networks learning}\label{sec:training}

We now derive a scheme for learning noisy-OR BNs that uses parallel max-product for fast approximate inference.

\subsection{Deriving the Elbo}\label{sec:deriving} 
Noisy-OR BNs are directed models with intractable likelihood. Therefore, a standard approach is to maximize the evidence lower bound (Elbo) \citep{kingma2013auto}:
\begin{align}\label{elbo}
\log p(x | \Theta) 
&\ge \mathbb{E}_{q(h|x, \phi)}\left\{\log p(h, x | \Theta) - \log q(h|x, \phi) \right\} \nonumber \\
&= \mathbb{E}_{q(h|x, \phi)}\left\{\log p(h, x | \Theta) \right\} 
+  \mathbb{H} \left\{ q(h |x, \phi) \right\} \nonumber \\
&= \mathcal{L}(x, \Theta, \phi),
\end{align}
where $q(h|x, \phi)$ is an approximate posterior, which VI assumes to be the output of a recognition network \citep{buhai2020empirical} or a MF posterior \citep{jaakkola1999variational,ji2020variational}. The first term in Equation \eqref{elbo} is the expectation of the joint log-likelihood under the approximate posterior distribution, while the second term is the entropy of the approximate posterior.
If we set $q(h|x, \phi) = p(h|x, \Theta)$, then the bound in Equation \eqref{elbo} becomes tight. However, the exact posterior of a noisy-OR BN is intractable.

We propose to derive an approximate posterior for a binary observation $x$ as follows. We first use max-product to either \textbf{(a)} estimate the mode of the model posterior $\tilde{h}(x, T=0) \approx \argmax p(h|x, \Theta)$ or \textbf{(b)} get a sample from the model posterior $\tilde{h}(x, T=1) \sim p(h|x, \Theta)$. Similar to \citet{lazaro2021perturb}, we address these posterior queries by clamping the visible variables to their observed value and running max-product, i.e., we set $\lambda_i=(0, -\infty)$ if $x_i=0$, $\lambda_i=(-\infty, 0)$ if $x_i=1$ in Problem \eqref{pmap}. We then solve Problem \eqref{pmap} with a temperature $T=0$ for \textbf{(a)} and $T=1$ for \textbf{(b)} using the PMP method described in Section \ref{sec:pmp}. We refer to the posterior inference query \textbf{(a)} or \textbf{(b)} as:
\begin{equation}\label{h-from-pmp}
\tilde{h}(x, T)=\texttt{PMP}(x, ~ \Theta, ~ T).
\end{equation}
After addressing \textbf{(a)} or \textbf{(b)}, we define the approximate posterior $q(h|x)$ by a Dirac delta centered at $\tilde{h}(x,T)$: $q(h|x)=\mathbf{1}(h=\tilde{h}(x,T) )$.
The lower bound in Equation \eqref{elbo} becomes $\mathcal{L}(x, \Theta) = \log p(\tilde{h}(x, T), x ~|~ \Theta)$.
$\mathcal{L}$ does not depend on $\phi$, and the entropy of $q(h|x)$ is $0$. Let $z=(z_0, \tilde{h}(x,T), x)$. Equation \eqref{noisyOR} can then be used to decompose the Elbo as a sum over the different factors:
\begin{align}\label{elbo-bp}
\mathcal{L}(x, \Theta) = &\sum_{i=1}^{m+n} 
z_i \log\Bigl( 1 - \exp \Bigl(-\theta_{0 \to i} - \sum_{k \in \mathcal{P}(i)} \theta_{k \to i} z_k \Bigr) \Bigr) \nonumber \\
& + (1 - z_i) \Bigl(-\theta_{0 \to i} - \sum_{k \in \mathcal{P}(i)} \theta_{k \to i} z_k \Bigr).
\end{align}

\subsection{Optimizing the Elbo}\label{sec:optimizing} 
The Elbo in Equation \eqref{elbo-bp} admits a closed-form gradient. Let us denote $f(\beta) = \log(1 - \exp(-\beta))$, whose derivative is $f'(\beta) = \frac{\exp(-\beta)}{1 - \exp(-\beta)}$. Let $k \in \mathcal{P}(i)$. Then the partial derivative of the Elbo w.r.t. $\theta_{k \to i}$ is:
\begin{equation}\label{elbo-grad}
\frac{\partial \mathcal{L}(z, \Theta)}{\partial \theta_{k \to i}}
= z_i z_k f'\Bigl(\theta_{0 \to i} + \sum_{k \in \mathcal{P}(i)} \theta_{k \to i} z_k \Bigr) + (z_i - 1) z_k
\end{equation}
A similar relationship holds for  $\frac{\partial \mathcal{L}(z, \Theta)}{\partial \theta_{0 \to i}}$, by setting $z_0=1$. 

\textbf{Parameter sharing:} In Sections \ref{sec:BMF} and \ref{sec:BD}, several parent-child pairs $(k, i)$ of the noisy-OR BN use the same parameter $\theta$. The chain rule generalizes the partial derivative w.r.t. $\theta$ by summing the right-hand side of Equation \eqref{elbo-grad} over the pairs sharing this parameter.

\textbf{Stochastic gradients updates: } We iterate through the data via mini-batches  \citep{robbins1951stochastic}, and we form a noisy estimate of the gradient of the Elbo on each mini-batch. This allows (a) scalability of our approach to large datasets (b) escaping local optima. We then use \texttt{Adam} \citep{kingma2014adam} to update the parameters $\Theta$. Finally, as in \citet{ji2020variational}, we clip the parameters $\Theta = \max (\Theta, \epsilon)$ to keep the Elbo in Equation \eqref{elbo-bp} finite. Algorithm 1 summarizes one step of parameters updates.

\begin{algorithm}[!htbp]
   \caption{Stochastic gradient updates with max-product}
   \label{alg:example}
\begin{algorithmic}
   \STATE {\bfseries Input:} Current parameters $\Theta^{(t)}$
   \newline
   Current mini-batch $\mathcal{B}^{(t)}$ of size $S$
   \newline
   Max-product temperature $T$
   \newline
   Learning rate \texttt{lr}
   \newline
   Clipping value $\epsilon$
   \STATE {\bfseries Output:} Updated parameters $\Theta^{(t+1)}$
   \FUNCTION{\texttt{UpdateParameters}}
   \FOR{$x_i \in \mathcal{B}^{(t)}$}
   \STATE $\tilde{h}_i(x_i,T) =\texttt{PMP}(x_i, ~ \Theta^{(t)}, ~T)$ as in Equation \eqref{h-from-pmp}
   \STATE Compute $\nabla \mathcal{L}(x_i, \Theta^{(t)})$ using Equation \eqref{elbo-grad}
   \ENDFOR
   \STATE $\nabla \mathcal{L}_{\mathcal{B}^{(t)}}(\Theta^{(t)}) = \frac{1}{S} \sum_{x_i \in \mathcal{B}^{(t)}} \mathcal{L}(x_i, \Theta^{(t)})$
   \STATE $\Theta^{(t + 1)} = \texttt{ADAM}(\Theta^{(t)}, \nabla \mathcal{L}_{\mathcal{B}^{(t)}}(\Theta^{(t)}), \texttt{lr})$
   \STATE $\Theta^{(t + 1)} =\max(\Theta^{(t + 1)}, \epsilon)$
   \ENDFUNCTION
\end{algorithmic}
\end{algorithm}

\subsection{Robustifying VI using MP}\label{sec:bp+vi}

Our objective value differs from the one in \citet{ji2020variational}. Algorithm 1 optimizes the Elbo defined in Equation \eqref{elbo-bp}---referred to as $\text{Elbo}^{\MP}$---w.r.t. the model parameters for a given binary configuration---while \citet{ji2020variational} optimize an Elbo derived using MF VI---referred to as $\text{Elbo}^{\VI}$. When both are defined, $\text{Elbo}^{\MP}$ and $\text{Elbo}^{\VI}$ are two valid lower bounds of the log-likelihood of a noisy-OR BN. Thus, in the rest of this paper, we refer to the Elbo of a method as the maximum of $\text{Elbo}^{\VI}$ and $\text{Elbo}^{\MP}$---Appendix \ref{appendix:mode_estimation} discusses how we can also define $\text{Elbo}^{\MP}$ for any VI posterior.

When the approximate posterior is concentrated into a single Dirac delta, $\text{Elbo}^{\MP}$ is tighter than $\text{Elbo}^{\VI}$: $\text{Elbo}^{\VI}$ is derived from $\text{Elbo}^{\MP}$ using Jensen's inequality in \citet[Eq. (6)]{ji2020variational}---see Appendix \ref{sec:appendix_binary_posterior} for more details. However, the non-zero entropy term present in $\text{Elbo}^{\VI}$ makes it often tighter when the approximate posterior is not a Dirac delta. The optimization of $\text{Elbo}^{\VI}$ using the simplistic MF posterior is hard and often gets stuck in bad local optima. This explains the catastrophic failures of MF VI in Sections \ref{sec:BMF} and \ref{sec:BD}. In contrast, MP uses a richer posterior which makes the optimization of $\text{Elbo}^{\MP}$ easier. As a result, our approach seems better at parameter search. We then propose to robustify MF VI with a hybrid approach, which uses the parameters $\Theta^{\Alg1}$ learned with Algorithm 1 to initialize the VI training from \citet{ji2020variational}. This initialization should guide the parameter search of VI and lead to a better optima than standalone VI, while returning a tighter Elbo.

\section{Computational results}\label{sec:exp}
We assess the performance of our methods on five categories of binary datasets 
(a) the \texttt{tiny20} dataset discussed in \citet{ji2020variational} (b) five large sparse \texttt{Tensorflow} datasets, (c) binary matrix factorization datasets (d) seven synthetic datasets introduced in \citet{buhai2020empirical} (e) the 2D blind deconvolution dataset from \citet{lazaro2021perturb}. 

Each experiment is run on a NVIDIA Tesla P100.

\subsection{Methods compared}
We compare the following methods in our experiments:
\newline
$\bullet$ \textbf{Full VI}: this is the approach from \citet{ji2020variational}. The authors did not release their code. To efficiently use their method in our experiments, we re-implemented it in \texttt{JAX} \citep{jax2018github}, using the variational hyperparameters reported. We use \texttt{ADAM} \citep{kingma2013auto} as we observe that it leads to better performance than the preconditioning proposed by the authors.
\newline
$\bullet$ \textbf{Local VI}: This is our re-implementation of the local models proposed by \citet{ji2020variational} and described in Section \ref{sec:related}, which are required to scale VI to large sparse datasets.
\newline
$\bullet$ \textbf{MP}: this is the proposed max-product training described in Algorithm 1. Max-product is run with a damping $\alpha=0.5$ for $N_{\MP}=100$ iterations. We select the temperature $T\in \{0,1\}$ with better empirical performance.
\newline
$\bullet$ \textbf{MP + VI}: this is the hybrid training proposed in Section \ref{sec:bp+vi}. We first run Algorithm 1 to learn the parameters $\Theta^{\Alg1}$, then run VI training for a few iterations starting from $\Theta^{\Alg1}$.

All the methods consider a clipping value $\epsilon=10^{-5}$ for the parameters $\Theta$. For a given experiment, all the methods use the same learning rate and mini-batch size, and we report the best performance of each method over several initializations---which we describe in Appendix \ref{appendix:init}.

\subsection{Tiny20 dataset}\label{sec:tiny20}
\textbf{Dataset: } We first consider the \texttt{tiny20} dataset\footnote{Accessible at \href{https://cs.nyu.edu/~roweis/data/20news_w100.mat}{https://cs.nyu.edu/$\sim$roweis/data/20news\_w100.mat}} on which \citet{ji2020variational} illustrate many of their findings. As in \citet{ji2020variational}, we build a three-layers graph with $100$ visible and $44$ hidden nodes using the procedure in Appendix \ref{appendix:graph} and we train on $70\%$ of the data at random (i.e. $11,369$ samples).

\textbf{Training: } We train full VI and local VI for  $1,500$ gradient steps, and for $5,000$ steps.
For MP + VI, we first run $1,000$ gradient steps using Algorithm 1 with $T=0$, then $500$ gradient steps using VI. All the methods use full-batch gradients as in \citet{ji2020variational} and a learning rate of $0.01$. 

\begin{table}[!htbp]
\centering
\begin{tabular}{p{0.18\textwidth}p{0.12\textwidth} p{0.15\textwidth}}
    \toprule
    Method & Num iters & Test Elbo \\
    \toprule
    \toprule
    Full VI & $1.5$k & $- 14.41~(0.02)$  \\
    \midrule
    Full VI & $5$k & $- 14.40~(0.02)$ \\
    \midrule
    Local VI & $1.5$k & $- 14.43~(0.02)$ \\
    \midrule
    Local VI & $5$k & $- 14.43~(0.02)$ \\
    \midrule
    MP (ours) & $1$k &  $- 14.49~(0.03)$ \\
    \midrule
    MP + VI (ours) & $1.5$k &  $\mathbf{- 14.34}~(0.02)$ \\
    \bottomrule
\end{tabular}
\caption{Test Elbos on the \texttt{tiny20} dataset averaged over $10$ runs. Higher is better. Our hybrid method outperforms full and local VI.}
\label{table:tiny20}
\end{table}

\textbf{Results: } Table \ref{table:tiny20} reports the test Elbo (defined in Section \ref{sec:bp+vi} as the best value between $\text{Elbo}^{\VI}$ and $\text{Elbo}^{\MP}$) of the different methods averaged over $10$ random train-test splits. Our hybrid MP + VI approach outperforms all the variational methods by a statistically significant margin. Interestingly, we observe that (a) increasing the number of training iterations slightly improves full and local VI, but it does not make them competitive with our best method; (b) standalone MP is competitive; and (c) as reported in \citet{ji2020variational}, full VI performs slightly better than local VI, as the latter optimizes a constrained VI objective. 

In addition, we note that \citet{ji2020variational} reported a lower Elbo of $-14.50$ for their best full VI method, using $145$ nodes (as we do) with a different graph heuristic and a different initialization procedure---both not described. Finally, to illustrate the distinction between $\text{Elbo}^{\MP}$ and $\text{Elbo}^{\VI}$, we report these two metrics in Appendix \ref{appendix:elbo_mode_tiny20}, Table \ref{table:appendix_tiny20}. In particular, standalone MP is the best performer for $\text{Elbo}^{\MP}$.

\subsection{Large sparse Tensorflow text datasets}\label{sec:tensorflow}
\textbf{Dataset: } We compare our hybrid method with \citet{ji2020variational} on five large sparse \texttt{Tensorflow} text datasets \citep{tensorflow2015-whitepaper}, which respectively contain scientific documents, news, movie reviews, patent descriptions and Yelp reviews. Note that \citet{ji2020variational} only consider two datasets and do not detail their processing procedure. To process each dataset, we first tokenize and vectorize it using a vocabulary size of $10,000$ (removing all the words outside the vocabulary) and a maximum sequence length of $500$. Second, we represent each sentence by a binary vector $x\in \{0, 1\}^{10,000}$, where $x_j=1$ if the $j$th word is present. Our datasets' statistics are summarized in Appendix \ref{sec:tensorflow-statistics}, Table \ref{table:tensorflow-statistics}. Finally, as in the large sparse experiments of \citet{ji2020variational}, we build a five-layers graph for each dataset.

\textbf{Training: } We train local VI for $4,000$ gradient steps. For hybrid training, we use $3,600$ steps of Algorithm 1 with $T=0$, then $400$ steps of local VI training. Both methods use a mini-batch size of $128$ and a learning rate of $3 \times 10^{-4}$.

\begin{table}[!b]
\centering
\begin{tabular}{p{0.15\textwidth}p{0.2\textwidth} p{0.2\textwidth}}
    \toprule
    Dataset & Local VI & Hybrid (ours) \\
    \toprule
    \toprule
    \small{\texttt{Abstract}} & $-327.19~(0.05)$ & $-\mathbf{324.79}~(0.05)$ \\
    \midrule
    \small{\texttt{Agnews}} & $-130.90~(0.07)$ & $-\mathbf{126.48}~(0.02)$ \\
    \midrule
    \small{\texttt{IMDB}} & $-429.54~(0.02)$  & $-\mathbf{428.40}~(0.01)$ \\
    \midrule
    \small{\texttt{Patent}} & $-578.41~(0.04)$ & $-\mathbf{578.33}~(0.02)$ \\
    \midrule
    \small{\texttt{Yelp}} & $-294.46~(0.16)$ & $-\mathbf{292.08}~(0.02)$ \\
    \bottomrule
\end{tabular}
\caption{Test Elbos on the large sparse \texttt{Tensorflow} text datasets averaged over $10$ runs.}
\label{table:tfds}
\end{table}

\textbf{Results: } Table \ref{table:tfds} averages the test Elbo of both methods over $10$ runs---each run shuffles the training and test set separately. Our hybrid method outperforms local VI on four datasets and is tied on one. 
In addition, Table \ref{table:appendix_tensorflow} in Appendix \ref{appendix:elbo_mode_tensorflow} compares $\text{Elbo}^{\MP}$ with $\text{Elbo}^{\VI}$ on each dataset: the hybrid approach is the best performer for $\text{Elbo}^{\MP}$ on all the datasets, which shows that hybrid training improves the overall performance of the noisy-OR models.

\textbf{Timings comparison: } Table \ref{table:tfds_time} in Appendix \ref{sec:tensorflow-time} reports the update times (defined as the average time for one gradient step) of MP and local VI on each dataset: MP is two to four times faster. Despite updating the variational parameters one by one, local VI runs at a reasonable speed as it uses small arrays to represent large sparse datasets. Note that MP runs in parallel and does not exploit the sparsity of the data.

\subsection{Binary Matrix Factorization}\label{sec:BMF}
\textbf{Problem: } Our next problem is Binary Matrix Factorization (BMF). Let $n, r, p$ be three integers with $r < \min(n, p)$ and let $U \in  \{0, 1\}^{n \times r}, V \in  \{0, 1\}^{r \times p}$ be two binary matrices. We assume that addition is performed on the Boolean semi-ring, i.e. $1+1=1$, and we define a binary matrix $X= U V \in  \{0, 1\}^{n \times p}$. The BMF problem consists in recovering the binary matrices $U$ and $V$ given the observations $X$. 

This problem is equivalent to learning a noisy-OR BN with $p$ visible nodes and $r$ hidden nodes, and with the parameters $\theta^x, \theta^u \in \mathbb{R}_+, \hat{V} \in \mathbb{R}_+^{r \times p}$, such that (a) the failure probability between the $i$th hidden and the $j$th visible variable is given by $\exp(-\hat{V}_{ij})$ (b) the prior probability of each hidden variable is equal to $1 - \exp(-\theta^u)$ (c) the noise probability of each visible variable is $1 - \exp(-\theta^x)$. Note that $\theta^x$ (resp. $\theta^u$) is shared across all the visible (resp. hidden) variables.  Let $\Theta=(\theta^x, \theta^u,\hat{V} )$. For $x \in \{0,1\}^p$, the conditional probability of the $j$th entry $x_j$ is
\begin{equation*}
p(x_j=1 ~|~ u_1, \ldots, u_r, \Theta) = 1 - \exp \Big( -\theta^x - \sum_{i=1}^r \hat{V}_{ij}u_i \Big).
\end{equation*}
The rows of $X$ give access to $n$ such observations, and our Algorithm 1 naturally extends to the BMF problem. 

\textbf{Dataset: } We fix $n=p$ and consider two increasing sequences of values for $n \in \{100, 200, 400\}$ and for $r/n \in \{0.2, 0.4, 0.6\}$. We additionally fix the probability $p_X = p(X_{ij}=1)=0.25, ~\forall i,j$. To do this, we first set  $p_{UV} = p(U_{ik}=1)=p(V_{kj}=1) = \sqrt{1 - (1 - p_X)^{1 / r}}, \forall i,j,k$. We then generate three matrices $V \in  \{0, 1\}^{r \times p}, U^{\text{train}} \in  \{0, 1\}^{n \times r}, U^{\text{test}} \in  \{0, 1\}^{n \times r}$ with prior $p_{UV}$ and define  $X^{\text{train}} = U^{\text{train}} V$, $X^{\text{test}} = U^{\text{test}} V$.

\textbf{Related work: } \citet{ravanbakhsh2016boolean} proposed to learn $U$ and $V$ with max-product by estimating the mode of the joint posterior $\max_{U,V}p(U,V|X)$, using non-symmetric priors for $U$ and $V$. Their method is very similar to PMP \citep{lazaro2021perturb} which proposes to sample from the joint multimodal posterior to solve the 2D blind deconvolution problem, Section \ref{sec:BD}. Both approaches do not consider training and directly solve max-product inference, which cannot be expressed in a mini-batch format and has to run on all the training data simultaneously. These two methods are then memory-intensive, and cannot scale to datasets orders of magnitude larger than the ones used here. In comparison, our MP approach computes the gradient of $\text{Elbo}^{\MP}$ for each training sample, which is memory-light and allows scaling to larger datasets. We report the results of PMP here, which we accelerate on GPU with \texttt{PGMax} \citep{zhou2022pgmax}, and we use $p_{UV}$ as priors for $U$ and $V$.

\textbf{Training: } We train full VI and BP for $40,000$ gradient steps with batch size $20$ and learning rate $0.001$. We use MP with $T=1$ to sample from the posterior as it allows to escape local optima during training. For PMP, there is no training and we directly turn to inference using $1,000$ max-product iterations as in \citet{lazaro2021perturb}.

\textbf{Metrics: } We report the Elbo of each method, as well as its update time. We also report its test reconstruction error, which is defined as $\frac{1}{n^2} \| U^{\text{test}} ~ \hat{V}^{\text{thre}} - X^{\text{test}} \|_1$, where $U^{\text{test}}$ and $\hat{V}^{\text{thre}}$ are binary matrices and have used $1+1=1$ for multiplication. $\hat{V}^{\text{thre}}$ is derived by thresholding the learned $\hat{V}$ with a threshold of $\log(2)$: a $1$ in $\hat{V}^{\text{thre}}$ corresponds to a failure probability lower than $0.5$ in $\hat{V}$. $U^{\text{test}}$ is the mode of posterior, estimated as detailed in Appendix \ref{appendix:mode_estimation}. For PMP, $\hat{V}$ is already binary and we only report its test RE---the update times are not defined for PMP as there is no training.

\begin{table*}[!t]
\centering
\resizebox{\textwidth}{!}{
\begin{tabular}{p{0.04\textwidth}p{0.04\textwidth}p{0.17\textwidth}p{0.15\textwidth}p{0.17\textwidth} p{0.17\textwidth}p{0.15\textwidth}p{0.17\textwidth}p{0.15\textwidth}}
\toprule
\multicolumn{2}{c}{Dataset} & \multicolumn{3}{c}{Full VI} & \multicolumn{3}{c}{MP (ours)} & \multicolumn{1}{c}{PMP} \\
\cmidrule(lr){1-2}
\cmidrule(lr){3-5}
\cmidrule(lr){6-8}
\cmidrule(lr){9-9}
$n$ & $r$ & Test Elbo $\uparrow$ & Test RE $(\%)$ $\downarrow$   & Update time (s) $\downarrow$  & Test Elbo $\uparrow$ & Test RE $(\%)$ $\downarrow$ & Update time (s) $\downarrow$ & Test RE $(\%)$ $\downarrow$ \\
\toprule
\toprule
$100$ & $20$ & $-18.26~(0.76)$ & $\mathbf{4.32}~(0.51)$ & $0.62~(0.03)$ & $\mathbf{-17.01}~(1.28)$ & $4.44~(0.64)$ & $\mathbf{0.09}~(0.00)$ & $9.81~(0.92)$\\

& $40$ & $-43.80~(3.27)$ & $\mathbf{9.15}~(1.42)$ & $1.39~(0.05)$ & $\mathbf{-39.29}~(0.90)$ & $9.75~(0.30)$ & $\mathbf{0.11}~(0.00)$ & $9.63~(0.67)$\\

& $60$ & $-78.42~(3.18)$ & $10.93~(0.98)$ & $2.56~(0.07)$ & $\mathbf{-54.25}~(1.24)$ & $13.68~(0.53)$ & $\mathbf{0.13}~(0.00)$ & $\mathbf{7.36}~(1.03)$\\
\midrule
$200$ & $40$ & $-103.03~(17.31)$ & $10.80~(2.07)$ & $2.48~(0.05)$ & $\mathbf{-42.71}~(2.07)$ & $\mathbf{7.05}~(0.49)$ & $\mathbf{0.14}~(0.00)$ & $12.57~(0.52)$\\

& $80$ & $-295.83~(2.95)$ & $24.98~(0.32)$ & $6.74~(0.07)$ & $\mathbf{-80.23}~(1.68)$ & $11.71~(0.40)$ & $\mathbf{0.20}~(0.14)$ & $\mathbf{11.05}~(0.56)$\\

& $120$ & $-362.32~(5.78)$ & $24.55~(0.34)$ & $14.42~(0.42)$ & $\mathbf{-95.25}~(1.46)$ & $13.11~(0.35)$ & $\mathbf{0.26}~(0.15)$ & $\mathbf{8.11}~(0.68)$\\
\midrule
$400$ & $80$ & ~~~~~~~~--- & ~~~~~~~~--- & $18.49~(0.06)$ & $\mathbf{-94.95}~(2.19)$ & $\mathbf{9.72}~(0.26)$ & $\mathbf{0.35}~(0.00)$ & $12.95~(0.83)$\\

& $160$ &  ~~~~~~~~--- & ~~~~~~~~--- & $72.22~(0.30)$ & $\mathbf{-152.16}~(3.21)$ & $\mathbf{11.06}~(0.26)$ & $\mathbf{0.61}~(0.00)$ & $11.74~(0.47)$\\

& $240$ & ~~~~~~~~--- & ~~~~~~~~--- & $167.83~(0.20)$ & $\mathbf{-154.94}~(1.71)$ & $\mathbf{10.82}~(0.27)$ & $\mathbf{0.86}~(0.00)$ & ~~~~~~~~---\\
\bottomrule
\end{tabular}}
\caption{BMF results averaged over $10$ runs. Arrows pointing up (down) indicate that higher (lower) is better. For large settings, ``---'' means that we were not able to get the results, due to time-out (for VI) or out-of-GPU-memory error (for PMP).}
\label{table:BMF}
\end{table*}

\textbf{Results: } Table \ref{table:BMF} averages the results over $10$ runs---each run generate new $V, U^{\text{train}}, U^{\text{test}}$. For $n=100$, there is no clear winner: MP achieves a higher Elbo, while PMP and VI reach lower REs. However, the performance of VI decreases as $n$ increases: for $n=200, r\in\{80,120\}$, VI has a test RE very close to $p_X=0.25$, which is what would return the trivial estimate $\hat{V}=0$. In addition, the sequential MF parameters updates make full VI prohibitively slow here: MP is $55$ times faster for $n=200, r=120$, and $195$ times faster for $n=400, r=240$. Finally, for $n=400$, VI did not finish training after three weeks and the test REs are close to $p_X$, which shows that no latent structure has been recovered. PMP is a solid competitor: it is faster than our method as it has no learning, and it leads to better performance when $r/n=0.6$. However it cannot scale and runs out of GPU memory for the large $n=400, r=240$.

\subsection{Overparametrization experiments}\label{sec:overparam}
\textbf{Problem:} Here, we reproduce the noisy-OR experiment from \citet{buhai2020empirical}. The authors introduced seven synthetic datasets\footnote{All the datasets are at \footnotesize{\href{https://github.com/clinicalml/overparam/tree/master/noisyor/datasets}{https://github.com/clinicalml/overparam} }}---which we refer to as OVPM. Five datasets (\texttt{IMG}, \texttt{PLNT}, \texttt{UNIF}, \texttt{CON8}, \texttt{CON24}) are generated from ground truth (GT) noisy-OR BNs while two (\texttt{IMG-FLIP} and \texttt{IMG-UNIF}) additionally perturb the observations.

The five GT noisy-OR BNs have the same structure, defined as follows. $K^*=8$ latent variables $u_1, \ldots, u_{8}$, are each associated with a continuous vector of parameters $V^k \in \mathbb{R}_+^p$. $V^1, \ldots,V^{8}$ are shared across all the observations while $u_k$ expresses whether the $k$th latent variable is active for a given observation. Each latent variable has a prior $1-\exp(-\theta_k)$ with $\theta_k\ge0$. Let $\Theta^*=(V^1, \ldots,V^8, \theta_1, \ldots, \theta_8, \theta^x)$. An observation $x$ is generated such that
$$
p(x_{j}=1 ~|~ u_1, \ldots, u_r, \Theta^*) = 1 - \exp \Big( -\theta^x - \sum_{k=1}^8 u_k V^k_{j} \Big), ~ \forall j.
$$
The GT parameters $\Theta^*$ are different for each GT noisy-OR BN. Figure \ref{fig:ovpm_bd}[left] shows $V^1, \ldots,V^{8}$ and eight cluttered binary samples from one of the datasets, \texttt{IMG}\footnote{\texttt{IMG} is originally from \citet{vsingliar2006noisy}.}.

\textbf{Training: } \citet{buhai2020empirical} learned the noisy-OR BN above for increasing values $K \ge 8$ of latent variables and study how overparametrization improves the recovery of the GT parameters $V^1, \ldots,V^{8}$. We compare our MP approach with their results, using $T=1$ for MP. For each dataset, we then consider an increasing sequence of latent variables $K\in\{8, 10, 16, 32\}$. We use the same training parameters as \citet{buhai2020empirical}: $9,000$ training samples, $100$ epochs, a batch size of $20$ and a learning rate of $0.001$.

\textbf{Metrics: } We compare the performance of our method with the VI results reported in the main Figure 2 of \citet[Fig. 2]{buhai2020empirical} (the numerical values are in Tables 2 and 5 in their appendices). As the authors, we report the averaged number of GT parameters $V^1, \ldots,V^{8}$ recovered during training---which we compute as detailed in Appendix \ref{appendix:ovpm_metric}---and the percentage of runs with full recovery.

\begin{figure*}[!t]
    \centering
    \includegraphics[width=\textwidth]{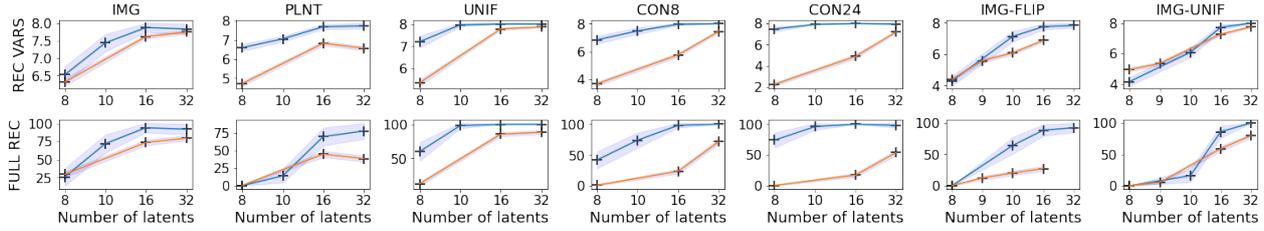}
    \caption{Results for the seven OVPM datasets (a) averaged over $50$ repetitions for our method (blue) and (b) reported in \citet{buhai2020empirical} for VI (orange). As in \citet{buhai2020empirical}, we report the $95\%$ confidence intervals (CIs): the authors considered $500$ repetitions, which explain their smaller CIs. Black markers indicate the number of latent variables for which each method is evaluated: \citet{buhai2020empirical} did not evaluate VI for $K=10$ on the first five datasets.
    [Top] Averaged number of GT parameters recovered. 
    [Bottom] Percentage of runs where all the GT parameters are recovered. For overparametrized models with $16$ or $32$ latent variables, our method always recovers a higher number of GT parameters. All the numerical values are reported in Appendix \ref{appendix:ovpm_table}, Table \ref{table:appendix_ovpm}.}
    \label{fig:overparam}
\end{figure*}

\textbf{Results: } 
Figure \ref{fig:overparam} compares our method (blue) averaged over $50$ repetitions, with VI (orange). Both MP and VI benefit from overparametrization: when $K$ increases, both methods recover more GT parameters. In addition, MP outperforms VI. In particular, for each dataset, for a model with $16$ or $32$ latent variables, our method (a) always recovers on average at least seven (out of eight) GT parameters (b) always performs better than VI. This gap is larger for the first five datasets, which do not perturb the observations.

\subsection{2D Blind Deconvolution}\label{sec:BD} 

\begin{figure*}[!t]
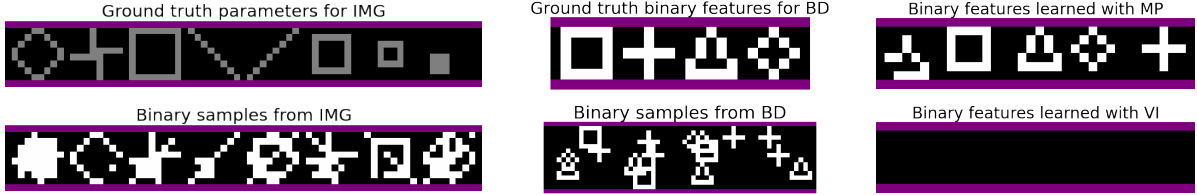

    \centering
    \begin{tabular}{c}
		\includegraphics[height=0.0525\textheight]{IMG_W.pdf}\\
		\includegraphics[height=0.0525\textheight]{IMG_X.pdf}\\
	\end{tabular}
    \begin{tabular}{c}
		\includegraphics[height=0.0525\textheight]{W.pdf}\\
		\includegraphics[height=0.0525\textheight]{X_small.pdf}
	\end{tabular}
    \begin{tabular}{c}
		\includegraphics[height=0.0525\textheight]{very_small_learned_mp_binary.pdf}\\
		\includegraphics[height=0.0525\textheight]{very_small_learned_vi_binary.pdf}
	\end{tabular}
    \caption{[Top left] Continuous GT parameters for the \texttt{IMG} dataset. Grey (resp. black) pixels correspond to failure probabilities of $0.1$ (resp. $1.0$). [Bottom left] $8$ cluttered samples from the \texttt{IMG} dataset. [Top middle] GT binary features for the BD problem. [Bottom middle] $4$ samples from the BD dataset.
    [Right] Binary features learned with MP [top] and VI [bottom] for BD for a random seed.
    }
    \label{fig:ovpm_bd}
\end{figure*}

\textbf{Problem: } Our last experiment is the 2D blind deconvolution (BD) problem from \citet[Section 5.6]{lazaro2021perturb}. The task consists in recovering two binary variables $W$ and $S$ from $100$ binary images\footnote{To generate the datasets, we use the publicly released code at \href{https://github.com/vicariousinc/perturb_and_max_product/blob/master/experiments/generate_convor.py}{https://github.com/vicariousinc/perturb\_and\_max\_product} } $X\in \{0,1\}^{n \times p}$. $W$ (size: $n_\text{feat}\times \text{feat}_\text{height}\times \text{feat}_\text{width}$) contains 2D binary features. $S$ (size: $n_\text{images}\times n_\text{feat} \times \text{act}_\text{height}\times \text{act}_\text{width} $) is a set of binary indicator variables. $S$ and $W$ are combined by convolution, placing the features defined by $W$ at the locations specified by $S$ in order to form $X$. Unlike $S$, $W$ is shared by all images. The dimensions of the GT $W$ used to generate $X$ are $4\times 5\times 5$, but the authors set the dimensions of the learned $\hat{W}$ to $5\times 6\times 6$, which we do too. Figure \ref{fig:ovpm_bd}[center] shows the ground truth $W$ and four samples from $X$ from the BD dataset. Appendix \ref{appendix:BD_simple} presents another example from \citet{lazaro2021perturb}, which visualizes $S,W$ and $X$ on a simpler dataset.

The BMF experiment, Section \ref{sec:overparam}, is a particular case of BD: BD is a harder problem. BD is also equivalent to learning a noisy-OR BN, which we describe in Appendix \ref{appendix:BD_noisyOR}.

\textbf{Methods compared: }
We compare full VI and MP with PMP (discussed in Section \ref{sec:BMF}), which directly learns a binary $\hat{W}$ by sampling from the joint posterior $p(S, W | X)$. 

\textbf{Training: } We train MP and full VI for $3,000$ steps on $80\%$ of the data, using full-batch gradients and a learning rate of $0.01$. PMP has no training and, for inference, we use the same priors as by \citet{lazaro2021perturb}

\textbf{Metrics: } We report the test Elbo, the update time and the test RE of each method. Here, the test RE is defined as $\frac{1}{np} \| X_{\text{RE}}^{\text{test}} - X^{\text{test}} \|_1$, where $X_{\text{RE}}^{\text{test}}$ is computed by convolving the estimated test feature locations $S^{\text{test}}$ with the thresholded learned features $\hat{W}^{\text{thre}}$. Finally, we match $\hat{W}^{\text{thre}}$ with the GT $W$ using intersection over union (IOU) for matching and report the features IOU---defined in Appendix \ref{appendix:BD_iou}. For PMP, we only report the test RE and features IOU.

\textbf{Results: } Table \ref{table:BD} averages the four metrics over $10$ repetitions with random train-test splits. VI is $50$ times slower than our method and completely fails at recovering the latent structure of the data, which leads to worse test metrics. Again, PMP is a strong competitor. However, (a) it is sensitive to the value of the priors of $X$, $S$ and $W$, (b) it leads to a test RE twice higher than MP, (c) it is memory-intensive. 

Figure \ref{fig:ovpm_bd}[right] shows the binary $\hat{W}^{\text{thre}}$ learned with MP and VI for a random seed---all the results are in Appendix \ref{appendix:BD_features}. MP recovers the four GT features and adds a noisier feature in the first position (which has a smaller prior and can be easily discarded) while VI fails. Finally, we refer to Appendix \ref{appendix:BD_rec} for a comparison of the reconstructed test images returned by our method and PMP.

\begin{table}[!b]
\centering
\begin{tabular}{p{0.2\textwidth}p{0.18\textwidth} p{0.18\textwidth}p{0.18\textwidth}}
    \toprule
    Metrics & Full VI & PMP & MP (ours) \\
    \toprule
    \toprule
    Test RE $(\%)$ $\downarrow$ & $23.46~(0.50) $ & $6.65~(0.79)$ & $\mathbf{2.96}~(0.23)$ \\
    \midrule
    Features IOU $\uparrow$ & $0.00~(0.00)$ & $0.94~(0.03)$ & $\mathbf{0.99}~(0.01)$ \\
    \midrule
    Test Elbo $\uparrow$ & $-233.22~(1.33)$ & N/A & $\mathbf{-43.12} (1.36)$ \\
    \midrule
    Update time (s) $\downarrow$ & $24.65~(0.67)$ & N/A & $\mathbf{0.53} (0.00)$ \\
    \bottomrule
\end{tabular}
\caption{BD results averaged over $10$ runs. Arrows pointing up (down) indicate that higher (lower) is better. VI fails while our method recovers all the features. PMP is a solid competitor.}
\label{table:BD}
\end{table}

\section{Discussion}
We have developed a fast, memory-efficient, stochastic algorithm for training noisy-OR BNs. Contrary to existing VI approaches with a recognition network, our MP method induces explaining-away and recovers more GT parameters on the OVPM datasets. In contrast with MF VI approaches, our method (a) finds better local optima; and (b) scales to large dense datasets. This explains, respectively, why (a) it solves the BD and the BMF problems while MF VI catastrophically fails; and (b) it is up to two orders of magnitude slower. Finally, our method is more memory-efficient than PMP. In addition, we have proposed to use our method to guide VI and help it find better local optima on the large sparse real \texttt{Tensorflow} datasets. Our next line of work is to use our algorithm to train noisy-OR BNs on complex synthetic scenes and extract rich latent representations.

\bibliographystyle{plainnat}
\bibliography{arxiv}

\newpage
\appendix
\onecolumn
\section{An equivalent representation of a noisy-OR factor}\label{appendix:pgmax-noisy-or}

We discuss herein an equivalent representation of the noisy-OR conditional distribution in Equation \eqref{noisyOR} that uses the tractable factors supported by \texttt{PGMax}.

First, as we have observed in Section \ref{sec:max-product}, for the messages from factors to variables, the max-product updates detailed in Equations \eqref{eq:bpupdates} require to loop through all the valid configurations of a factor. Let $i$ be a variable in the graph, let $N_i=| \mathcal{P}(i)|$ be the cardinality of the set $\mathcal{P}(i)$ of parents of $i$, and let $\mathcal{P}(i)=\{ j_1, \ldots, j_{N_i} \}$. The noisy-OR factor associated with $i$, and described in Equation \eqref{noisyOR}, connects the $2 + N_i$ variables $\{z_i\} \cup\{z_0\} \cup z_{\mathcal{P}(i)}$, and the has $2^{2 + N_i}$ valid configuration. Consequently, a naive implementation of the max-product message updates in Equations \eqref{eq:bpupdates} has a complexity exponential in the number of variables of the noisy-OR factors, which is prohibitively slow for large factors.

To remedy this, let us introduce a family of ``noise-free'' OR factors which are described by the conditional distribution
\begin{align}\label{noisefreeOR}
p(z_i=0 ~|~ z_{\mathcal{P}(i)}) 
= \prod_{k \in \mathcal{P}(i)} (1 - z_k).
\end{align}
The noise-free OR factors simply express the logical condition $z_i = \text{OR} (z_{j_1}, \ldots, z_{j_{N_i}})$. They do not involve the noisy-OR parameters $\Theta$ defined in Equation \eqref{noisyOR}. 

It turns out that, for a ``noise-free'' OR factor, the messages updates derived in \texttt{PGMax} have a complexity linear in the number of variables connected to this factor. Consequently, if we derive an equivalent representation of the noisy-OR conditional distribution in Equation \eqref{noisyOR} that uses the noise-free OR conditional distribution in Equation \eqref{noisefreeOR}, the cost of the max-product messages updates (using \texttt{PGMax}) would go from $\mathcal{O}(2^{| \mathcal{P}(i)|})$ down to $\mathcal{O}(| \mathcal{P}(i)|)$.

\begin{figure*}[!htbp]
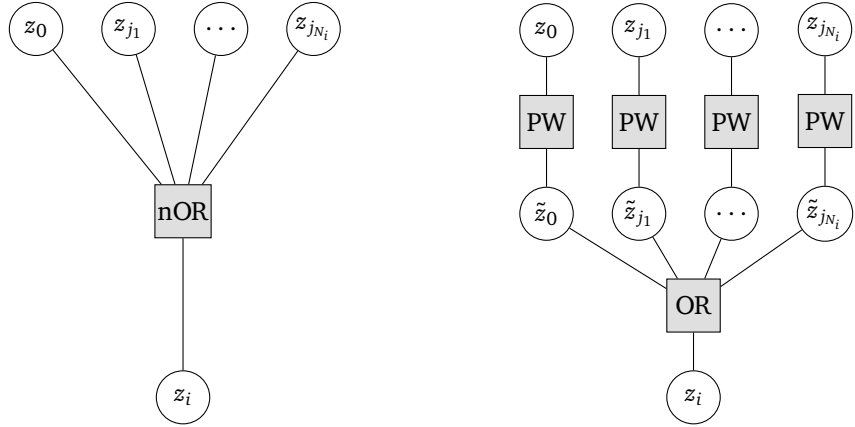

\centering
\tikz{ 
    \node[latent] (zi) {$z_i$}; 
    \node[obs, rectangle, above=1.75 of zi] (nOR) {nOR}; 
    \node[latent, above=1.7 of nOR, xshift=5mm] (z2) {$\ldots$}; 
    \node[latent, above=1.7 of nOR, right=.5 of z2] (z3) {$z_{j_{N_i}}$}; 
    \node[latent, above=1.7 of nOR, left=.5 of z2] (z1) {$z_{j_1}$}; 
    \node[latent, above=1.7 of nOR, left=.5 of z1] (z0) {$z_0$}; 
    \factoredge[-] {z0,z1,z2,z3} {nOR} {zi};
    
    \node[latent, right=6 of zi] (ztildei) {$z_i$}; 
    \node[obs, rectangle, above=.5 of ztildei] (OR) {OR}; 
    \node[latent, above=.5 of OR, xshift=5mm] (ztilde2) {$\ldots$}; 
    \node[latent, above=.5 of OR, right=.5 of ztilde2] (ztilde3) {$\tilde{z}_{j_{N_i}}$}; 
    \node[latent, above=.5 of OR, left=.5 of ztilde2] (ztilde1) {$\tilde{z}_{j_1}$}; 
    \node[latent, above=.5 of OR, left=.5 of ztilde1] (ztilde0) {$\tilde{z}_0$}; 
    \factoredge[-] {ztilde0,ztilde1,ztilde2,ztilde3} {OR} {ztildei};

    \node[obs, rectangle, above=.5 of ztilde2] (PW2) {PW}; 
    \node[obs, rectangle, above=.5 of ztilde3] (PW3) {PW}; 
    \node[obs, rectangle, above=.5 of ztilde1] (PW1) {PW}; 
    \node[obs, rectangle, above=.5 of ztilde0] (PW0) {PW}; 
    \node[latent, above=.5 of PW2] (zbis2) {$\ldots$}; 
    \node[latent, above=.5 of PW3] (zbis3) {$z_{j_{N_i}}$}; 
    \node[latent, above=.5 of PW1] (zbis1) {$z_{j_1}$}; 
    \node[latent, above=.5 of PW0] (zbis0) {$z_0$}; 
    \factoredge[-] {zbis2} {PW2} {ztilde2};
    \factoredge[-] {zbis3} {PW3} {ztilde3};
    \factoredge[-] {zbis1} {PW1} {ztilde1};
    \factoredge[-] {zbis0} {PW0} {ztilde0};
}
\caption{[Left] Noisy-OR factor graph with conditional distribution given by Equation \eqref{noisyOR}. [Right] Equivalent factor graph, which involves a noise-free OR factor with conditional distribution given by Equation \eqref{noisefreeOR} and several pairwise factors. For the latter, \texttt{PGMax} can perform GPU-accelerated messages updates with a complexity linear in the number of variables connected to the factor.}
\label{fig:pgmax-noisyOR}
\end{figure*}

To this end, we define two factor graphs, which we both represent in Figure \ref{fig:pgmax-noisyOR}:
\begin{enumerate}
\item  The first factor graph considers a single noisy-OR factor---nOR in Figure \ref{fig:pgmax-noisyOR}---which connects the leak variable $z_0$ and the parents variables $z_{\mathcal{P}(i)}$ to the child variable $z_i$ via the noisy-OR conditional distribution defined in Equation \eqref{noisyOR}.
\item The second factor graph introduces the auxiliary binary variables $\tilde{z}_0, \tilde{z}_{j_1}, \ldots, \tilde{z_{j_{N_i}}}$ and connects them to the child variable $z_i$ via the noise-free OR factor defined in Equation \eqref{noisefreeOR}. In addition, for each $k \in \{0\} \cup \mathcal{P}(i)$, there is a pairwise factor---referred to as PW in Figure \ref{fig:pgmax-noisyOR}---that connects the variables $z_k$ and $\tilde{z}_k$ and that is defined by
\begin{equation*}
\left\{
\begin{aligned}
p(\tilde{z}_k=0 ~|~ z_k=0) &= 1 \\
p(\tilde{z}_k=0 ~|~ z_k=1) &= \exp(- \theta_{k \to i}) \\
\end{aligned}
\right.
\end{equation*}
which can be represented in a more compact form by
\begin{equation}\label{eq:noise-free}
p(\tilde{z}_k=0 ~|~ z_k) = \exp(- \theta_{k \to i} z_k).
\end{equation}
\end{enumerate}

We aim at showing the equivalence between the two factor graphs. To this end, let us derive the conditional distribution of $z_i$ given $z_{\mathcal{P}(i)}$ for the second factor graph:
\begin{align*}
p(z_i=0 ~|~ z_{\mathcal{P}(i)}) 
&= \sum_{\tilde{z}_0 \in \{0,1\}, ~ \tilde{z}_{\mathcal{P}(i)} \in \{0, 1\}^{N_i} } p(z_i=0, ~\tilde{z}_0, ~ \tilde{z}_{\mathcal{P}(i)} ~|~ z_{\mathcal{P}(i)})  \\
&= \sum_{\tilde{z}_0 \in \{0,1\}, ~ \tilde{z}_{\mathcal{P}(i)} \in \{0, 1\}^{N_i} } 
p(z_i=0  ~|~ \tilde{z}_0, \tilde{z}_{\mathcal{P}(i)}) ~ p(\tilde{z}_0, \tilde{z}_{\mathcal{P}(i)} ~|~ z_{\mathcal{P}(i)}) \text{ by conditional independence} \\
&= \sum_{\tilde{z}_0 \in \{0,1\}, ~ \tilde{z}_{\mathcal{P}(i)} \in \{0, 1\}^{N_i} } ~
\prod_{k \in \{0\} \cup \mathcal{P}(i)} (1 - \tilde{z}_k) ~ p(\tilde{z}_k ~|~ z_k)\\
&= \prod_{k \in \{0\} \cup \mathcal{P}(i)} p(\tilde{z}_k=0 ~|~ z_k) \text{ as the product cancels if any $\tilde{z}_{k} =1$}\\
&= \exp (-\theta_{0 \to i}) \prod_{k \in \mathcal{P}(i)} \exp (- \theta_{k \to i} z_k) ~\text{using Equation \eqref{eq:noise-free} and $z_0=1$},
\end{align*}
which is exactly the noisy-OR conditional distribution Equation \eqref{noisyOR}. This proves the equivalence between the two factor graphs. In particular, we can use the second factor graph to represent a noisy-OR factor and benefit from the GPU-accelerated messages updates from \texttt{PGMax} which have a complexity linear in the number of variables.

\section{Graph generation procedure for multi-layered noisy-OR Bayesian networks}\label{appendix:graph}

We describe below the graph generation procedure we use to build the multi-layered noisy-OR BNs in the \texttt{tiny20} and the \texttt{Tensorflow} experiments, Sections \ref{sec:tiny20} and \ref{sec:tensorflow}.

We assume we are given a binary matrix $X \in \{0,1\}^{n \times p}$, where each row is a binary observation: as in Section \ref{sec:noisy-or} there are $p$ visible variables.  In the case of the \texttt{tiny20} and \texttt{Tensorflow} datasets, each visible variable corresponds to a word, and each observation to a sentence or a document: $X_{ij}=1$ indicates that the $j$th word is present in the $i$th document.

Given an integer $n_{\text{layers}}$, we aim at building a noisy-OR Bayesian network with $n_{\text{layers}} + 1$ layers. The bottom layer (with index $n_{\text{layers}}$) of the network contains all the visible nodes, while the trivial top layer (with index $0$) only contains the leak node. Our procedure builds the graph iteratively, from the top to the bottom by repeating the two following steps (for $j$ running from $n_{\text{layers}}$ down to $1$)
\begin{enumerate}
\item Build a distance matrix for the $j$th layer.
\item Create the variables of the $j-1$th layer and add edges connecting the parents of the $j-1$th layer to the children of the $j$th layer.
\end{enumerate}

We detail these two steps further below.

\paragraph{Distance matrix for the bottom layer: }
We first detail how we build the distance matrix for the bottom layer---which contains all the visible variables. We start by building a vector of empirical word frequencies $C \in \{0,1\}^{p}$ such that
$$
C_{j} = \frac{1}{N} \sum_{i=1}^n X_{ij}, ~ \forall j,
$$
is the empirical probability that the $j$th word appears in a document. We also define a matrix of empirical co-occurrence frequencies $O \in \{0,1\}^{p \times p}$ where
$$
O_{jk} = \frac{1}{N} \sum_{i=1}^n X_{ij} X_{ik}, ~ \forall j,k,
$$
is the empirical probability that the $j$th and $k$th words co-occur in a document. From there, we can define the empirical ratio 
$$
R_{jk} = \frac{O_{jk}}{C_j C_k}.
$$
$R_{jk}$ possesses a few interesting properties. First, from the law of large numbers, when $n$ grows to infinity, $C_j \to p(x_j=1)$, $O_{jk} \to p(x_j=1, x_k=1)$ and consequently $R_{jk} \to \frac{p(x_j=1, x_k=1)}{p(x_j=1) p(x_k=1)}$. Therefore, if the $j$th and $k$th words are independent, the limit of $R_{jk}$ in the case of an infinite amount of data is $1$. If the limit of $R_{jk}$ is higher than $1$, then $p(x_j=1, x_k=1)$ is higher than the case where the variables are independent. Finally, $R_{jk}$ can also be connected with the mutual information, commonly used in information theory---see \citet{globerson2004euclidean}. 

Given these properties, we propose to define the distance matrix associated with the bottom layer by
$$D^{(n_{\text{layers}})}_{jk} = \exp(-R_{jk}), ~ \forall j,k.$$

\paragraph{Building the $j-1$th layer and connecting it to the $j$th layer: }
Let $j \ge 2$. We assume that the $j$th layer has $d$ variables and that we are given a distance matrix $D^{(j)} \in \mathbb{R}_+^{d \times d}$---we have described above how to build $D^{(n_{\text{layers}})}$ for the bottom layer which has $p$ variables. We now describe how our procedure builds the $j-1$th layer and adds edges between the $j$th and $j-1$th layers. To this end, we use two hyperparameters: (a) the ratio between the number of variables of the $j$th layer and of the $j-1$th layer, $r_{\text{children to parents}}$ (which we set to $3$ in our experiments) (b) the number of nodes of the $j-1$th layer that each node of the $j$th layer will be connected with, $n_{\text{parents by node}}$ (which we set to $5$).

As a first step, we use hierarchical clustering\footnote{We use the \texttt{AgglomerativeClustering} procedure from \texttt{scikit-learn} \citep{scikit-learn}.} with average linkage on the distance matrix $D^{(j)}$ to form  $\lfloor \frac{d}{r_{\text{children to parents}}} \rfloor$ clusters, with indices $1, \ldots, \lfloor \frac{d}{r_{\text{children to parents}}} \rfloor$. For each cluster $m$, we then create a variable for the $j-1$th layer, $z_m^{(j-1)}$.

We refer to $\text{label}(z^{(j)}_k)$ as the label returned by this clustering step for the $k$th variable $z^{(j)}_k$ of the $j$th layer. We could then add edges between the $j$th and $j-1$th layers by going through the pairs of variables $\Big(z^{(j)}_{\ell}, z^{(j-1)}_{\text{label}(z^{(j)}_{\ell} )} \Big)_{\ell}$. However, if we were doing so, each variable of the $j$th layer would only be connected to one variable of the $j-1$th layer. The resulting noisy-OR BN would not be able to induce explaining-away (see Section \ref{sec:noisy-or}) as each effect would be connected to a single cause. To allow inference to induce this appealing property, we propose to add extra edges to the graph by connecting each variable of the $j$th layer to multiple variables of the $j-1$th layer as follows. First, we define the distance from a variable $z^{(j)}_k$ of the $j$th layer to a variable $z_m^{(j-1)}$ of the $j-1$th layer as the average distance from $z^{(j)}_k$ to all the elements of the $j$th layer with label $m$:
$$
\text{dist}(z^{(j)}_k, z_m^{(j-1)}) 
= \frac{1}{\left| \left\{  \ell:~\text{label}(z^{(j)}_{\ell})=m \right\} \right|} 
\sum_{\ell: ~ \text{label}(z^{(j)}_{\ell})=m} D^{(j)}_{k \ell}, ~ \forall k, m.
$$
Second, we add an edge connecting $z^{(j)}_k$ to the $n_{\text{parents by node}}$ variables of the $j-1$th layer with smallest $\text{dist}(z^{(j)}_k, z_m^{(j-1)})$: this intuitively connects $z^{(j)}_k$ to the $n_{\text{parents by node}}$ labels it is the ``closest''. Each variable of the $j$th layer is now connected to the same number of variables of the $j-1$th layer above. However, each variable of the $j-1$th layer may be connected to a different number of variables of the $j$th layer: we denote $\mathcal{C}(z^{(j-1)}_{m})$ the set of indices of the variables of the $j$th layer connected to $z^{(j-1)}_{m}$. Let us note that, by definition, each node of the $j-1$th and $j$th layer is also connected to the leak node.

Our last step is to define the symmetric distance matrix $D^{(j-1)}$ between two variables of the $j-1$th layer, which we set to the average distance of all the variables of $j$th layer connected to these two variables:
$$
D_{m_1, m_2}^{(j-1)} 
= \frac{1}{\left| \mathcal{C}(z^{(j-1)}_{m_1}) \right|
~ \left| \mathcal{C}(z^{(j-1)}_{m_2}) \right|} 
\sum_{k \in \mathcal{C}(z^{(j-1)}_{m_1})} 
\sum_{\ell \in \mathcal{C}(z^{(j-1)}_{m_2})} 
D^{(j)}_{k \ell},
~\forall m_1, m_2.
$$

\textbf{Case $j$=1: } When $j=1$, as the $0$th layer only consists of the leak node, we simply connect each node of the first layer to it.

\paragraph{A comment for the tiny20 graph: } We mentioned that, for the \texttt{tiny20} experiment, our graph contains $145$ nodes and three layers (excluding the top layer). Our graph can be indeed decomposed as follows. The bottom layer contains $100$ visible nodes, the second layer contains $\lfloor \frac{100}{3} \rfloor=33$ hidden nodes, the first layer contains $\lfloor \frac{33}{3} \rfloor=11$ hidden nodes and the top layer only contains the leak node.

\section{Initialization procedures}\label{appendix:init}
This section describes the initialization procedures used in the different experiments.

\subsection{Tiny20 and large Tensorflow experiments}
For each method used in the \texttt{tiny20} and the \texttt{Tensorflow} experiments, Sections \ref{sec:tiny20} and \ref{sec:tensorflow}, we consider the four following initializations for the failure, prior, and noise, probabilities:
\begin{enumerate}
\item all the failure probabilities, all the prior probabilities and all the noise probabilities are set to $0.5$.
\item all the failure probabilities are set to $0.5$, all the prior and noise probabilities are set to $0.1$.
\item all the failure probabilities are set to $0.9$, all the prior and noise probabilities are set to $0.1$.
\item all the failure probabilities are set to $0.9$, all the prior and noise probabilities are set to $0.5$,
\end{enumerate}
Once we have initialized the aforementioned probabilities, we initialize the parameters $\Theta$ accordingly by using the fact that, that for a node $i$ and a node $k \in \mathcal{P}(i)$, the failure probability is $\exp(-\theta_{k\to i})$, while the noise probability---or prior probability when $\mathcal{P}(i)$ is empty---is $p(z_i=1 ~|~ z_{\mathcal{P}(i)}=0, ~\Theta) = 1 - \exp(-\theta_{0\to i})$.

For a given method and a given dataset, we run each initialization for $10$ different seeds. We then report the results for the initialization that leads to the best averaged test results.

\subsection{BMF and BD experiments}\label{sec:init-BMF}
For each method used in the BMF and BD experiments, Sections \ref{sec:BMF} and Sections \ref{sec:BD}, we initialize all the noise probabilities to $0.01$ and keep them fixed during training. We have found this to be particularly useful to avoid a local minima where (a) the noise probabilities converge to the average number of activations of the visible variables (b) the prior probabilities converge to $0$. 

We consider the four following initializations of the remaining failure and prior probabilities:
\begin{enumerate}
\item all the failure probabilities and all the prior probabilities are set to $0.5$.
\item all the failure probabilities are set to $0.5$, all the prior probabilities are set to $0.1$.
\item all the failure probabilities are set to $0.9$, all the prior probabilities are set to $0.1$.
\item all the failure probabilities are set to $0.9$, all the prior probabilities are set to $0.5$,
\end{enumerate}
In addition, the solution to the BMF and to the BP problems are invariant to certain permutations. For instance, a solution to the BMF problem is invariant to applying the same permutation on the columns of $U$ and the rows of $V$, while a solution to the BD problem is invariant to applying the same permutation on the features indices (the first dimension) of both $W$ and $S$. A uniform initialization would then induce symmetries in the parameters during training. To break these symmetries, we add some centered Gaussian noise $\mathcal{N}(0, 0.1)$ to the failure and prior probabilities, before projecting them to $[0, 1]$. 

As before, after initializing the failure and prior probabilities (and adding the Gaussian noise), we initialize the parameters $\Theta$ accordingly. For a given method and experiment, we run each experiment for $10$ different seeds and report the initialization that leads to the best averaged test results.

\subsection{OVPM experiment}
For the overparametrization experiment, Section \ref{sec:overparam}, we follow a very similar procedure to Section \ref{sec:init-BMF}, but we only consider the initialization methods 3 and 4, and run each initialization for $50$ seeds.

\section{Estimating the mode of the model posterior after inference}\label{appendix:mode_estimation}

Given a test sample $x$, we discuss how to estimate the mode of the model posterior $h^{\MAP} \approx \argmax p(h|x, \Theta)$ when we use VI and MP at inference time. We use this posterior mode estimation in our experiments to compute (a) $\text{Elbo}^{\MP}$ in the \texttt{tiny20} and the \texttt{Tensorflow} experiments, Sections \ref{sec:tiny20} and \ref{sec:tensorflow}, and (b) the test reconstruction errors in the BMF and BD experiments, Sections \ref{sec:BMF} and \ref{sec:BD}.

For MP, we estimate $h^{\MAP}$ by clamping the visible variables to their observed value and running max-product with a temperature $T=0$. This is exactly the inference query \textbf{(b)} discussed in Section \ref{sec:training}.

For VI, the inference from \citet{ji2020variational} gives access to the mean-field posterior parameters, that is, the parameters such that, the approximate posterior distribution factorizes as $q(h|x) = \prod_{i \in \mathcal{H}} q_i^{h_i} (1 - q_i)^{1 - h_i}$. We then estimate the mode of the posterior element-wise via rounding: $h^{\MAP}_i = \mathbf{1}(q_i \ge 0.5), ~\forall i$.

\section{Performance comparisons of  $\text{Elbo}^{\text{MP}}$ and  $\text{Elbo}^{\text{VI}}$}\label{appendix:elbo_mode}

This section compares $\text{Elbo}^{\MP}$ with $\text{Elbo}^{\VI}$ for the methods evaluated in the \texttt{tiny20} and the \texttt{Tensorflow} experiments, Sections \ref{sec:tiny20} and \ref{sec:tensorflow}.
\begin{enumerate}
\item $\text{Elbo}^{\VI}$ is computed by running the inference algorithm of \citet{ji2020variational}---which we have reimplemented.\\
\item  To compute $\text{Elbo}^{\MP}$, we estimate the posterior mode $h^{\MAP}$ as detailed in Section \ref{appendix:mode_estimation}, then plug it into Equation \eqref{elbo-bp}.
\end{enumerate}

\subsection{For a binary posterior, $\text{Elbo}^{\text{VI}}$ is a lower-bound of $\text{Elbo}^{\text{MP}}$}\label{sec:appendix_binary_posterior}

Let us briefly start by presenting the proof of a claim we made in Section \ref{sec:bp+vi}. We said that, for a binary observation $x \in \{0,1\}^p$, if the posterior $\tilde{h}(x, T)$ is binary, then $\text{Elbo}^{\VI}$ is a lower-bound of $\text{Elbo}^{\MP}$. To prove this point, let us assume that $\tilde{h}(x, T)$ is binary, let us introduce $z=(z_0,\tilde{h}(x, T),x)$ and let us recall that $\text{Elbo}^{\MP}$ is defined in Equation \eqref{elbo-bp} as follows:
\begin{align}\label{elbo-bp-appendix}
\begin{split}
\mathcal{L}(x, \Theta) 
&= \sum_{i=1}^{m+n} z_i \log\Bigl( 1 - \exp \Bigl(-\theta_{0 \to i} - \sum_{k \in \mathcal{P}(i)} \theta_{k \to i} z_k \Bigr) \Bigr) + (1 - z_i) \Bigl(-\theta_{0 \to i} - \sum_{k \in \mathcal{P}(i)} \theta_{k \to i} z_k \Bigr)\\
&= \sum_{i=1}^{m+n} z_i f\Bigl(\theta_{0 \to i} + \sum_{k \in \mathcal{P}(i)} \theta_{k \to i} z_k \Bigr) + (1 - z_i) \Bigl(-\theta_{0 \to i} - \sum_{k \in \mathcal{P}(i)} \theta_{k \to i} z_k \Bigr),
\end{split}
\end{align}
where we have used $f(\beta) = \log(1 - \exp(-\beta))$. Equation \eqref{elbo-bp-appendix} is exactly Equation (3) in \citet{ji2020variational} in the case of a binary posterior. From there, as $\theta_{0 \to i} \ge 0$ and $\theta_{k \to i} z_k \ge 0, ~ \forall k \in \mathcal{P}(i)$, the authors introduced an auxiliary parameter $r_{k\to i}$ for each edge connecting a non-leak parent variable to a child variable such that 
$$
r_{k\to i}\ge 0, ~ \forall k \in \mathcal{P}(i); \text{ and } \sum_{k \in \mathcal{P}(i)} r_{k\to i} = 1.
$$
Consequently $\sum_{k \in \mathcal{P}(i)} r_{k\to i} z_k \in [0,1]$. As $f$ is concave, the authors use Jensen's inequality to get the following lower-bound:
\begin{align}\label{elbo-bp-lower-bound}
\begin{split}
f\bigg(\theta_{0 \to i} + \sum_{k \in \mathcal{P}(i)} \theta_{k \to i} z_k \bigg)
&= f\bigg( \Bigl(1- \sum_{k \in \mathcal{P}(i)} r_{k\to i} z_k \Bigr) \theta_{0 \to i}
+ \sum_{k \in \mathcal{P}(i)} r_{k\to i} z_k \Bigl( \theta_{0 \to i} + \frac{\theta_{k \to i}}{r_{k \to i}} \Bigr) \bigg)\\
&\ge \Bigl(1- \sum_{k \in \mathcal{P}(i)} r_{k\to i} z_k \Bigr) f(\theta_{0 \to i})
+ \sum_{k \in \mathcal{P}(i)} r_{k\to i} z_k f(u_{k \to i}) 
\text{ where } u_{k \to i} = \theta_{0 \to i} + \frac{\theta_{k \to i}}{r_{k \to i}}\\
&= f(\theta_{0 \to i})
+ \sum_{k \in \mathcal{P}(i)} r_{k\to i} z_k \Bigl(f(u_{k \to i}) -  f(\theta_{0 \to i}) \Bigr)\\
\end{split}
\end{align}
By pairing Equations \eqref{elbo-bp-appendix} and \eqref{elbo-bp-lower-bound} we get:
\begin{equation}\label{elbo-vi}
\mathcal{L}(x, \Theta) 
\ge 
\sum_{i=1}^{m+n} z_i 
\left\{
f(\theta_{0 \to i})
+ \sum_{k \in \mathcal{P}(i)} r_{k\to i} z_k \Bigl(f(u_{k \to i}) -  f(\theta_{0 \to i}) \Bigr)
\right\}
+ (1 - z_i) \Bigl(-\theta_{0 \to i} - \sum_{k \in \mathcal{P}(i)} \theta_{k \to i} z_k \Bigr).
\end{equation}
The right-hand size of Equation \eqref{elbo-vi} is exactly $\text{Elbo}^{\VI}$ in the case of a binary posterior, as defined in \citet[Equation (9)]{ji2020variational}. Consequently, Equation \eqref{elbo-vi} proves that for a binary posterior, $\text{Elbo}^{\MP}$ is a tighter lower-bound of the intractable log-likelihood of a noisy-OR BN than $\text{Elbo}^{\VI}$. Hence, in all our experiments, we never compute $\text{Elbo}^{\VI}$ for a binary posterior.

\subsection{Performance comparisons on the tiny20 dataset}\label{appendix:elbo_mode_tiny20}

Table \ref{table:appendix_tiny20} reports the averaged test $\text{Elbo}^{\MP}$ and $\text{Elbo}^{\VI}$ on the \texttt{tiny20} dataset. Standalone MP is trained with Algorithm 1 to optimize $\text{Elbo}^{\MP}$. As a result, the MP parameters land in a local optima of this loss and MP reaches the highest test $\text{Elbo}^{\MP}$. MP is also the worst performer for $\text{Elbo}^{\VI}$ as it has not been exposed to this loss during training. 

In comparison, all the methods trained with $\text{Elbo}^{\VI}$ (including the hybrid method MP+VI) perform better at test time for $\text{Elbo}^{\VI}$ than for $\text{Elbo}^{\MP}$. Full VI performs particularly poorly for $\text{Elbo}^{\MP}$ as it has never been exposed to it during training. 

Finally, Table \ref{table:appendix_tiny20} suggests that initializing the VI training with MP helps VI find a better local optima of $\text{Elbo}^{\VI}$, which is why our hybrid method reaches the best overall lower bound---while maintaining a high $\text{Elbo}^{\MP}$.

\begin{table}[!ht]
\centering
\begin{tabular}{p{0.18\textwidth}p{0.1\textwidth} p{0.14\textwidth}p{0.14\textwidth}}
    \toprule
    Method & Num iters & Test $\text{Elbo}^{\MP}$ & Test $\text{Elbo}^{\VI}$ \\
    \toprule
    \toprule
    Full VI & $1.5$k & $-14.80~(0.03)$ & $- 14.41~(0.02)$  \\
    \midrule
    Full VI & $5$k & $-14.85~(0.03)$ & $- 14.40~(0.02)$ \\
    \midrule
    Local VI & $1.5$k & $-14.65~(0.03)$ & $- 14.43~(0.02)$ \\
    \midrule
    Local VI & $5$k & $-14.64~(0.03)$ & $- 14.43~(0.02)$ \\
    \midrule
    MP (ours) & $1$k & $\mathbf{-14.49}~(0.03)$  & $- 14.49~(0.03)$ \\
    \midrule
    MP + VI (ours) & $1.5$k & $-14.55 (0.02)$ & $\mathbf{- 14.34}~(0.02)$ \\
    \bottomrule
\end{tabular}
\caption{Test $\text{Elbo}^{\MP}$ and $\text{Elbo}^{\VI}$ on the \texttt{tiny20} dataset averaged over $10$ runs.}
\label{table:appendix_tiny20}
\end{table}

\subsection{Performance comparisons on the large sparse Tensorflow datasets}\label{appendix:elbo_mode_tensorflow}

Table \ref{table:appendix_tensorflow} reports the averaged $\text{Elbo}^{\MP}$ and $\text{Elbo}^{\VI}$ on the large sparse \texttt{Tensorflow} datasets. As before, standalone MP is the worst performer for $\text{Elbo}^{\VI}$ as it has not been exposed to this loss during training. Local VI is also the worst overall method for $\text{Elbo}^{\MP}$ for a similar reason. However, it performs better than MP on two datasets, which suggests that, for these datasets, standalone MP is stuck in a local optima during its training. 

Our hybrid method is the best performer for both $\text{Elbo}^{\MP}$ and $\text{Elbo}^{\VI}$, which shows that our MP approach finds a good area of the parameters space, that is further refined during the VI optimization of $\text{Elbo}^{\VI}$. As a result, the hybrid scheme improves the overall performance of each noisy-OR model.

\begin{table}[htbp!]
\centering 
\resizebox{\textwidth}{!}{
\begin{tabular}{
p{0.14\textwidth}
p{0.16\textwidth}p{0.16\textwidth}p{0.16\textwidth} p{0.01\textwidth}
p{0.16\textwidth}p{0.16\textwidth}p{0.16\textwidth}
}
    \cmidrule(lr){1-1}
    \cmidrule(lr){2-4}
    \cmidrule(lr){6-8}
    Dataset & Local VI, $\text{Elbo}^{\MP}$ & MP, $\text{Elbo}^{\MP}$ & Hybrid, $\text{Elbo}^{\MP}$ && Local VI, $\text{Elbo}^{\VI}$ & MP, $\text{Elbo}^{\VI}$ & Hybrid, $\text{Elbo}^{\VI}$ \\
    \toprule
    \toprule
    \texttt{Abstract} & $-342.79~(0.05)$ & $-342.48~(0.07)$ & $\mathbf{-327.73}~(0.06)$ && $-327.19~(0.05)$ & $-335.56~(0.05)$ & $\mathbf{-324.89}~(0.05)$ \\
    \midrule
    \texttt{Agnews} & $-134.98~(0.07)$ & $-140.48~(0.05)$ & $\mathbf{-127.75}~(0.02)$ && $-130.90~(0.07)$ & $-137.88~(0.08)$ & $\mathbf{-126.48}~(0.02)$  \\
    \midrule
    \texttt{IMDB} & $-450.48~(0.06)$ & $-438.16~(0.04)$ & $\mathbf{-431.34}~(0.03)$ && $-429.53~(0.02)$ & $-436.96~(0.04)$ & $\mathbf{-428.40}~0.01$  \\
    \midrule
    \texttt{Patent} & $-619.75~(0.07)$ & $-595.91~(0.10)$ & $\mathbf{-586.08}~(0.05)$ && $-578.41~(0.04)$ & $-590.33~(0.09)$ & $-\mathbf{578.33}~(0.07)$  \\
    \midrule
    \texttt{Yelp} & $-303.31~(0.07)$ & $-308.58~(0.09)$ & $\mathbf{-294.38}~(0.02)$ && $-294.16~(0.05)$ & $-302.75~(0.07)$ & $\mathbf{-292.08}~(0.02)$ \\
    \bottomrule
\end{tabular}
}
\caption{Test $\text{Elbo}^{\MP}$ and $\text{Elbo}^{\VI}$ on the large Tensorflow datasets averaged over $10$ runs.}
\label{table:appendix_tensorflow}
\end{table}

\section{Additional materials for the large sparse Tensorflow datasets}

This section reports some statistics for the large Tensorflow datasets used in Section \ref{sec:tensorflow}, as well as a timing comparison of the different methods used.

\subsection{Datasets statistics}\label{sec:tensorflow-statistics}

For the five large \texttt{Tensorflow} datasets, Table \ref{table:tensorflow-statistics} below gives access to (a) the full name of the dataset, as it appears in the catalog accessible at \href{https://www.tensorflow.org/datasets/catalog}{https://www.tensorflow.org/datasets/catalog} (b) the feature name used when loading the dataset (c) the number of edges in the BNs returned by our graph generation procedure (detailed in Appendix \ref{appendix:graph}) (d) the train and test set sizes. In particular, the BNs returned by our procedure have a similar number of edges. This is explained by the fact that, for all the datasets, we use the same number of visible variables---$10,000$---during the preprocessing, and the same hyperparameters during the BN generation.

\begin{table}[!htbp]
\centering
\resizebox{\textwidth}{!}{
\begin{tabular}{p{0.09\textwidth}p{0.25\textwidth}p{0.15\textwidth}p{0.17\textwidth}p{0.1\textwidth}p{0.1\textwidth}}
    \toprule
    Dataset & Full name & Feature name & Number of edges & Train set & Test set\\
    \toprule
    \toprule
    \small{\texttt{Abstract}} & \small{\texttt{scientific\_papers}} & \small{\texttt{abstract}}  & $90,554$ & $203,037$ & $6,440$\\
    \midrule
    \small{\texttt{Agnews}} & \small{\texttt{ag\_news\_subset}} & \small{\texttt{description}} & $89,508$ & $120,000$ & $7,600$\\
    \midrule
    \small{\texttt{IMDB}} & \small{\texttt{imdb\_reviews}} & \small{\texttt{text}} & $91,234$ & $25,000$ & $25,000$\\
    \midrule
    \small{\texttt{Patent}} & \small{\texttt{big\_patent/f}} & \small{\texttt{description}} & $90,606$ & $85,568$  & $4,754$\\
    \midrule
    \small{\texttt{Yelp}} & \small{\texttt{yelp\_polarity\_reviews}} & \small{\texttt{text}} & $91,111$ & $560,000$ & $38,000$\\
    \bottomrule
\end{tabular}}
\caption{\small{\texttt{Tensorflow}} datasets full names and statistics.}
\label{table:tensorflow-statistics}
\end{table}

\begin{table}[b!]
\centering
\begin{tabular}{p{0.14\textwidth} p{0.14\textwidth} p{0.14\textwidth}}
    \toprule
    Dataset & Local VI & MP (ours) \\
    \toprule
    \toprule
    \small{\texttt{Abstract}} & $17.71~(0.05)$ & $\mathbf{7.51}~(0.01)$ \\
    \midrule
    \small{\texttt{Agnews}} & $13.12~(0.07)$ & $\mathbf{7.39}~(0.00)$ \\
    \midrule
    \small{\texttt{IMDB}} & $32.52~(0.11)$  & $\mathbf{7.51}~(0.00)$ \\
    \midrule
    \small{\texttt{Patent}} & $18.72~(0.15)$ & $\mathbf{7.50}~(0.00)$ \\
    \midrule
    \small{\texttt{Yelp}} & $18.89~(0.04)$ & $\mathbf{7.49}~(0.00)$ \\
    \bottomrule
\end{tabular}
\caption{Update times, in seconds, for local VI and MP on the large sparse \texttt{Tensorflow} datasets averaged over $10$ runs.}
\label{table:tfds_time}
\end{table}

\subsection{Update times for local VI and MP}\label{sec:tensorflow-time}

Table \ref{table:tfds_time} reports the update time of local VI and MP on the \texttt{Tensorflow} datasets, which we have defined in Section \ref{sec:tensorflow} as the average time for one gradient step. The MP gradients updates detailed in Algorithm 1 run at a very similar speed on all the datasets. Indeed, the complexity of the messages updates is similar across the datasets as (a) as the different BNs have a similar number of edges (as seen in Table \ref{table:tensorflow-statistics}) (b) MP does not use exploit the sparsity of the data and represents each sentence by a vector $x \in \{0,1\}^{10,000}$.

In contrast, as explained in Section \ref{sec:related}, the local models in VI represent each sentence by its active visible variables and by their ancestors. We have set the number of active visible variables per sentence to be at most $500$, and in practice it can be lower---some datasets only have a few tenths of active variables on average. Consequently, local VI represents sparse data using arrays three orders of magnitudes smaller than MP. Hence, although local VI updates its variational parameters sequentially, it is reasonably fast. Nonetheless, its update time is dataset-specific and it is two to four times slower than MP.

\section{Additional material for the overparametrization experiment}\label{appendix:ovpm}

This section contains some additional materials for the overparametrization experiment presented in Section \ref{sec:overparam}. First, we discuss the method proposed in \citet{buhai2020empirical} to compute the number of GT parameters recovered during training. Second, we report the table of results associated with Figure \ref{fig:overparam}.

\subsection{Computing the number of ground truth parameters recovered}\label{appendix:ovpm_metric}

We consider a trained noisy-OR BN with $K\ge8$ latent variables and learned parameters $\hat{\Theta}=(\hat{V}^1, \ldots, \hat{V}^K, \hat{\theta}_1, \ldots, \hat{\theta}_K, \hat{\theta}^x)$. We follow the procedure of \citet{buhai2020empirical} to count the number of recovered GT parameters  $V^1, \ldots, V^8$---let us trivially note that are at most eight recovered GT parameters. 

First, we discard the $\hat{V}^k$ with a prior probability $1-\exp(-\hat{\theta}_k)$ lower than $0.02$. Second, we perform minimum cost bipartite matching between the non-discarded learned parameters and the GT ones $V^1, \ldots, V^8$, using the $\ell_{\text{inf}}$ norm as the matching cost. Finally, we count as recovered all the GT parameters with a matching cost lower than $1.0$.

\subsection{Table of results}\label{appendix:ovpm_table}

Table \ref{table:appendix_ovpm} reports the numerical results of the OVPM experiment which are displayed in Figure \ref{fig:overparam}, Section \ref{sec:overparam}. For VI, we take the numbers from Tables 2 and 5 in the appendices of \citet{buhai2020empirical}, which are averaged over $500$ repetitions. For MP, our results are averaged over $50$ seeds. As in \citet{buhai2020empirical}, we report the $95\%$ confidence intervals of each method.

\begin{table*}[!htbp]
\centering
\resizebox{\textwidth}{!}{
\begin{tabular}{p{0.10\textwidth}p{0.16\textwidth}p{0.22\textwidth}p{0.2\textwidth}p{0.22\textwidth} p{0.2\textwidth}}
\toprule
\multicolumn{2}{c}{Dataset} & \multicolumn{2}{c}{VI} & \multicolumn{2}{c}{MP (ours)} \\
\cmidrule(lr){1-2}
\cmidrule(lr){3-4}
\cmidrule(lr){5-6}
Name & Latent variables & Parameters recovered & Full recovery (\%) & Parameters recovered & Full recovery (\%) \\
\toprule
\toprule
\texttt{IMG} & $8$ & $6.31~(0.11)$ & $29.6~(4.0)$ & $\mathbf{6.52}~(0.24)$ & $\mathbf{26.0}~(12.1)$\\

& $10$ & N/A & N/A & $7.44~(0.25)$ & $72.0~(12.5)$\\

& $16$ & $7.62~(0.06)$ & $73.6~(3.9)$ & $\mathbf{7.88}~(0.13)$ & $\mathbf{94.0}~(6.0)$\\

& $32$ & $7.75~(0.05)$ & $79.6~(3.5)$ & $\mathbf{7.84}~(0.15)$ & $\mathbf{92.0}~(7.5)$\\
\midrule
\texttt{PLNT} & $8$ & $4.71~(0.12)$ & $\mathbf{0.4}~(0.6)$ & $\mathbf{6.60}~(0.18)$ & $0.0~(0.0)$\\

& $10$ & N/A & N/A & $7.06~(0.15)$ & $14.0~(9.6)$\\

& $16$ & $6.83~(0.12)$ & $45.0~(4.4)$ & $\mathbf{7.70}~(0.13)$ & $\mathbf{70.0}~(12.7)$\\

& $32$ & $6.57~(0.11)$ & $38.4~(4.3)$ & $\mathbf{7.74}~(0.15)$ & $\mathbf{78.0}~(11.5)$\\
\midrule
\texttt{UNIF} & $8$ & $5.35~(0.14)$ & $12.6~(2.9)$ & $\mathbf{7.20}~(0.27)$ & $\mathbf{60.0}~(13.6)$\\

& $10$ & N/A & N/A & $7.96~(0.08)$ & $98.0~(3.9)$\\

& $16$ & $7.78~(0.05)$ & $85.4~(3.1)$ & $\mathbf{8.00}~(0.00)$ & $\mathbf{100.0}~(0.0)$\\

& $32$ & $7.87~(0.04)$ & $88.2~(2.8)$ & $\mathbf{8.00}~(0.00)$ & $\mathbf{100.0}~(0.0)$\\
\midrule
\texttt{CON8} & $8$ & $3.70~(0.15)$ & $1.2~(1.0)$ & $\mathbf{6.84}~(0.27)$ & $\mathbf{42.0}~(13.7)$\\

& $10$ & N/A & N/A & $7.96~(0.08)$ & $98.0~(3.9)$\\

& $16$ & $5.77~(0.15)$ & $23.6~(3.7)$ & $\mathbf{7.96}~(0.06)$ & $\mathbf{98.0}~(3.9)$\\

& $32$ & $7.45~(0.08)$ & $71.6~(4.0)$ & $\mathbf{8.00}~(0.00)$ & $\mathbf{100.0}~(0.0)$\\
\midrule
\texttt{CON24} & $8$ & $2.26~(0.15)$ & $0.4~(0.6)$ & $\mathbf{7.48}~(0.24)$ & $\mathbf{74.0}~(12.2)$\\

& $10$ & N/A & N/A & $7.92~(0.00)$ & $96.0~(5.4)$\\

& $16$ & $4.90~(0.21)$ & $17.2~(3.3)$ & $\mathbf{8.00}~(0.00)$ & $\mathbf{100.0}~(0.0)$\\

& $32$ & $7.21~(0.10)$ & $53.8~(4.4)$ & $\mathbf{7.96}~(0.08)$ & $\mathbf{98.0}~(3.9)$\\
\midrule
\texttt{IMG-FLIP} & $8$ & $\mathbf{4.40}~(0.10)$ & $\mathbf{0.2}~(0.4)$ & $4.30~(0.32)$ & $0.0~(0.0)$\\

& $10$ & $6.09~(0.12)$ & $20.0~(3.5)$ & $\mathbf{7.14}~(0.33)$ & $\mathbf{64.0}~(13.3)$\\

& $16$ & $6.88~(0.09)$ & $27.0~(3.9)$ & $\mathbf{7.76}~(0.18)$ & $\mathbf{88.0}~(9.0)$\\

& $32$ & N/A & N/A & $7.84~(0.15)$ & $92.0~(13.5)$\\
\midrule
\texttt{IMG-UNIF} & $8$ & $\mathbf{4.95}~(0.12)$ & $0.0~(0.0)$ & $4.16~(0.28)$ & $0.0~(0.0)$\\

& $10$ & N/A & N/A & $6.08~(0.32)$ & $16.0~(12.2)$\\

& $16$ & $7.27~(0.09)$ & $59.0~(4.3)$ & $\mathbf{7.72}~(0.19)$ & $\mathbf{86.0}~(9.6)$\\

& $32$ & $7.76~(0.05)$ & $80.0~(3.5)$ & $\mathbf{8.00}~(0.00)$ & $\mathbf{100.0}~(0.0)$\\
\bottomrule
\end{tabular}}
\caption{Numerical results for the OVPM datasets. We use N/A to express that \citet{buhai2020empirical} did not evaluate VI for the associated number of latent variables. Our method outperforms VI on all the datasets. In particular, it always recovers more GT parameters in the overparametrized regime, that is for $16$ or $32$ latent variables. The performance gap is larger for the first five datasets (\texttt{IMG}, \texttt{PLNT}, \texttt{UNIF}, \texttt{CON8}, \texttt{CON24}) for which the data is not perturbed.}
\label{table:appendix_ovpm}
\end{table*}

\newpage
\section{Additional material for the 2D blind deconvolution experiment}\label{appendix:BD}

This section contains some additional materials for the 2D blind deconvolution (BD) experiment presented in Section \ref{sec:BD}. First, we discuss a simple example from \citet{lazaro2021perturb} which illustrates the generative process of the BD dataset. Second, we express the BD problem as a learning problem in a noisy-OR BN. Third, we define the features IOU metric used in Table \ref{table:BD}.  Finally, we display the continuous and binary features learned by each method, as well as the reconstructed test images for MP and PMP.

\subsection{A simple example}\label{appendix:BD_simple}

\begin{figure}[!htbp]
    \centering
    \begin{tabular}{c}
		\includegraphics[height=0.1\textwidth]{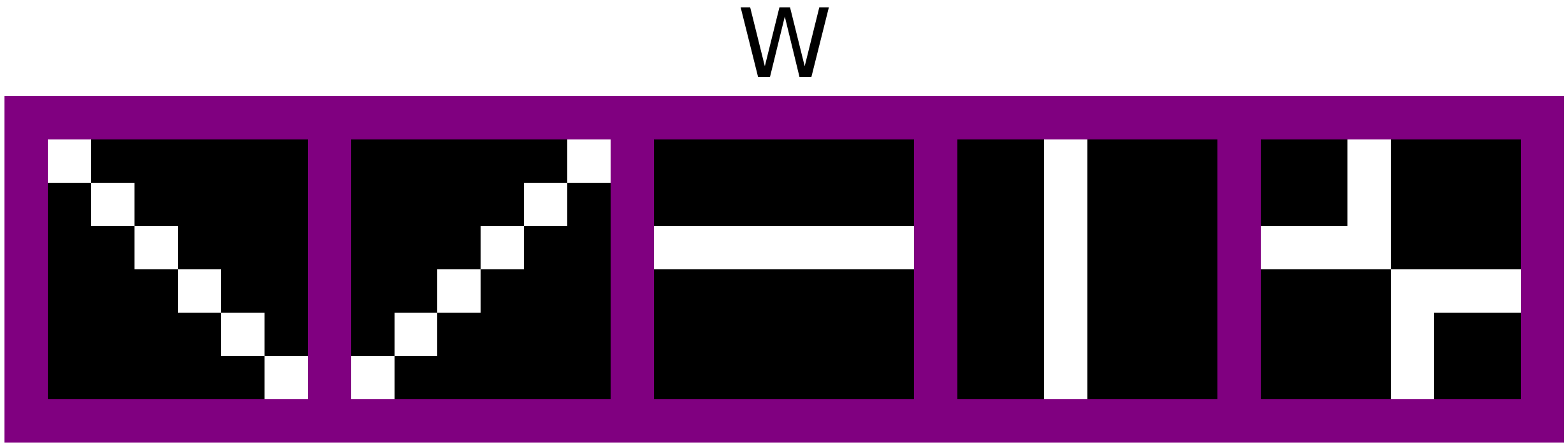}\\
		\includegraphics[height=0.15\textwidth]{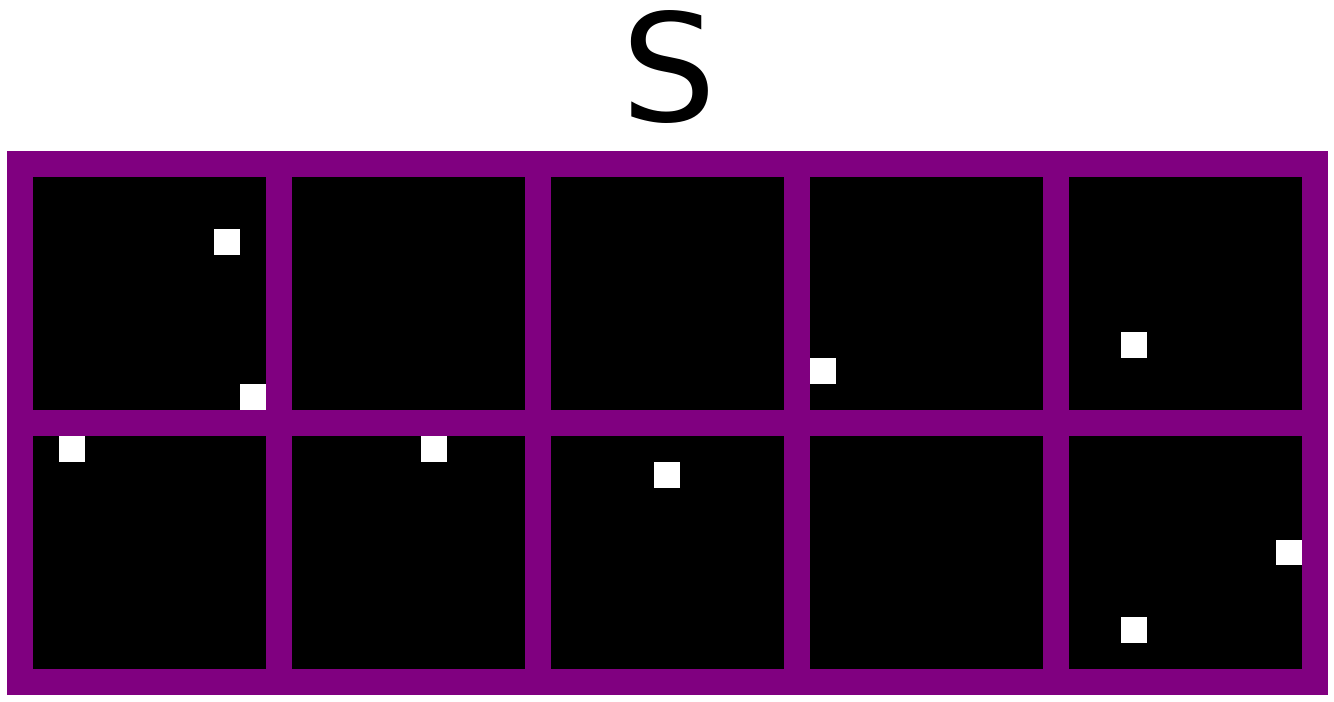} \includegraphics[height=0.15\textwidth]{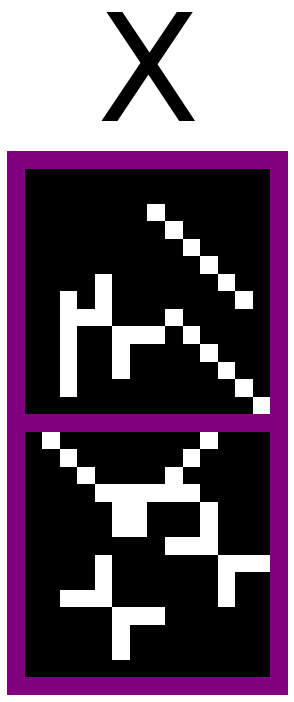}
	\end{tabular}
    \caption{Simple binary convolution example from \citet{lazaro2021perturb}, with features $W$, locations $S$ and resulting images $X$.}
    \label{fig:BD_simple}
\end{figure}

Figure \ref{fig:BD_simple} uses a simple example from \citet{lazaro2021perturb} to illustrate the generative process of the BD dataset. The small dataset considered here only contains two independent binary images: each image $X\in \{0,1\}^{15\times15}$ is formed by convolving the shared binary features $W\in \{0,1\}^{5\times6\times6}$ with the image-specific binary locations $S\in\{0,1\}^{5\times10\times10}$.

$W$ contains five features, each of size $6\times6$. $S$ contains the locations of the features, which are sampled at random using an independent Bernoulli prior per entry: $p(S_{f,i,j}=1)=0.01, ~\forall f,i,j$. The top (resp. bottom) row of $S$ indicates the locations of the features in the top (resp. bottom) image of $X$. The $j$th column of $S$ corresponds to the locations of the $j$th feature in $W$. For instance, the two activations on the right of the top-left block of $S$, means that the first feature in $W$ will appear twice on the right of the first image of $X$. This is verified by the two anti-diagonal lines in the top row of $X$.

\subsection{The BD problem can be expressed as learning a noisy-OR Bayesian network}\label{appendix:BD_noisyOR}
The 2D BD problem can be expressed as a learning problem in the noisy-OR BN detailed below.

Let $N\times P$ be the size of an image $X$. As $W$ is of size $n_\text{feat}\times \text{feat}_\text{height}\times \text{feat}_\text{width}$, $S$ is of size $n_\text{images}\times \text{act}_\text{height}\times \text{act}_\text{width}$, and $X$ is produced from $S$ and $W$ by convolution, let us first note that
\begin{align*}
N &= \text{act}_\text{height} + \text{feat}_\text{height} - 1\\
P &= \text{act}_\text{width} + \text{feat}_\text{width} - 1
\end{align*}
In addition, for a pixel with indices $(n, p)$, let us introduce the set of indices:
\begin{equation*}
\mathcal{I}(n,p) = \left\{ (i,j,k,\ell):
\begin{aligned}
& 1 \le i \le \text{act}_\text{height}\\
& 1 \le j \le \text{act}_\text{width}\\
& 1 \le k \le \text{feat}_\text{height}\\
& 1 \le \ell \le \text{feat}_\text{width}\\
& i + k - 1 = n\\
& j + \ell - 1 = p\\
\end{aligned}
\right\}.
\end{equation*}
The BD problem is equivalent to learning a noisy-OR BN where (a) the visible nodes are $X$ (b) the hidden nodes are $S$ (c) the positive continuous parameters are $\theta^x \in \mathbb{R}_+$, $\theta_1, \ldots, \theta_{n_\text{feat}} \in \mathbb{R}_+$, and $\hat{W}^\in\mathbb{R}_+^{ n_\text{feat}\times \text{feat}_\text{height}\times \text{feat}_\text{width} }$ and we denote $\Theta=(\theta^x, \theta_1, \ldots, \theta_{n_\text{feat}}, \hat{W})$ (d) the prior probability of each entry of S, for the $f$th feature $f$ is $p(S_{f,i,j}=1)=1 - \exp(-\theta_f), \forall i,j$ (e) the conditional probability of the pixel $X_{np}$ is given by
$$
p(X_{np}=1 ~|~ S, \Theta) = 1 - \exp \Bigg( -\theta^x - \sum_{1 \le f \le n_\text{feat}} \sum_{(i,j,k,\ell) \in \mathcal{I}(n,p)} S_{f,i,j} \hat{W}_{f,k,\ell} \Bigg)
$$
In particular, the noise probability of each visible variable is equal to $1 - \exp(-\theta^x)$.

\subsection{Computing the features intersection-over-union}\label{appendix:BD_iou}
Let us first define the intersection-over-union (IOU) between a thresholded learned feature $\hat{W}^{\text{thre}}_j \in \{0,1\}^{6 \times 6}$ and a GT feature $W_k\in \{0,1\}^{5 \times 5}$. To do so, we introduce the four sub-features $\hat{W}^{\text{thre}}_{j,1}, \ldots, \hat{W}^{\text{thre}}_{j,4}\in \{0,1\}^{5 \times 5}$ of the same size as $W_k$, obtained by removing the first or last row, and the first or last column of $\hat{W}^{\text{thre}}_j$. We then compute:
$$
\text{IOU}(\hat{W}^{\text{thre}}_j, W_k)
= \max_{\ell=1,\ldots,4}
\left\{
\frac
{\sum \limits_{1\le n,p\le5}
\text{AND}\Big( \big(\hat{W}^{\text{thre}}_{j, \ell} \big)_{np}=1, \big(W_k \big)_{np}=1 \Big)
}
{\sum \limits_{1\le n,p\le5}
\text{OR}\Big( \big(\hat{W}^{\text{thre}}_{j,\ell} \big)_{np}=1, \big(W_k \big)_{np}=1 \Big)
}
\right\}.
$$
The IOU is always between $0$ and $1$: an IOU of $0$ means that $\hat{W}^{\text{thre}}_j=0$ whereas an IOU of one means that one of the sub-features $\hat{W}^{\text{thre}}_{j,1}, \ldots, \hat{W}^{\text{thre}}_{j,4}$ is equal to $W_k$.

After training our noisy-OR BN on the BD problem, we perform minimum bipartite matching between the learned binary features $\hat{W}^{\text{thre}}_1, \ldots, \hat{W}^{\text{thre}}_5$ and the GT binary features $W_1, \ldots, W_4$, using the opposite of the IOU as the matching cost---as we want to maximize the IOU. We then define the features IOU as the average matching cost: a feature IOU of $1.0$ means that we have recovered the four GT features whereas an IOU of $0$ means that training has not learned any information.

\subsection{Learned binary features}\label{appendix:BD_features}

Our next Figure \ref{fig:appendix_BD1} plots the five continuous parameters $\hat{W}$ and the corresponding binary features $\hat{W}^{\text{thre}}$ learned by MP, VI, and PMP for each of the $10$ seeds. Note that the order of the features is not relevant here, as it depends on the random noise added to the unaries of each model during the initialization---as discussed in Appendix \ref{appendix:pgmax-noisy-or}. 

\begin{figure*}[!htbp]
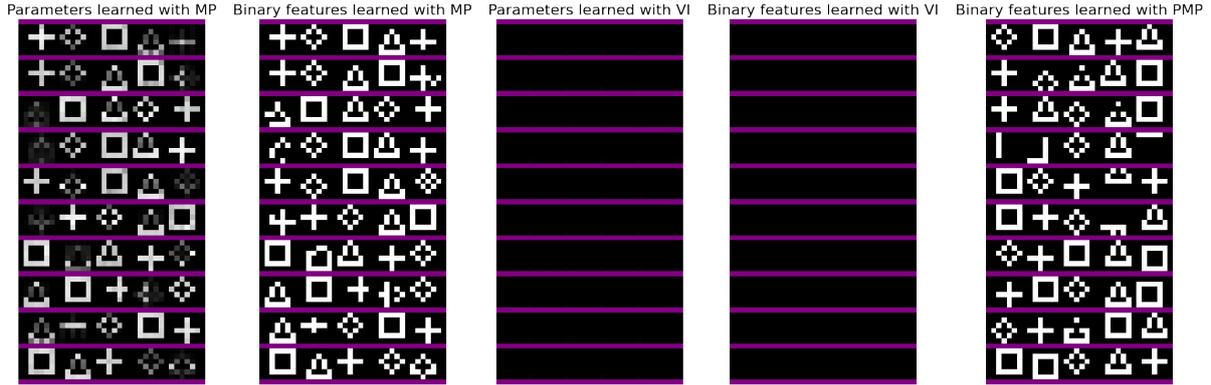

     \centering
     \includegraphics[height=0.31\textwidth]{learned_mp_continuous.pdf}
     \includegraphics[height=0.31\textwidth]{learned_mp_binary.pdf}
     \includegraphics[height=0.31\textwidth]{learned_vi_continuous.pdf}
     \includegraphics[height=0.31\textwidth]{learned_vi_binary.pdf}
     \includegraphics[height=0.31\textwidth]{PMP_learned_binary.pdf}
     \caption{
     [First panel] Continuous $\hat{W}$ learned with MP. [Second panel] Binary $\hat{W}^{\text{thre}}$ learned with MP. [Third panel] Continuous $\hat{W}$ learned with VI. [Fourth panel] Binary $\hat{W}^{\text{thre}}$ learned with VI. [Fifth panel] Binary $\hat{W}$ learned with PMP.}
    \label{fig:appendix_BD1}
\end{figure*}
VI completely fails at this task and does not learn any features. As PMP directly turns to posterior inference, the learned features $\hat{W}$ are binary so we only have one plot. PMP perfectly recovers the four GT features $W$ for seven of the ten runs. It misses two features on one run, and only misses one pixel of one feature on two runs. As the learned $\hat{W}^{\text{thre}}$ contains five features while the GT $W$ only contains four features, each run also learns an extra feature. However, PMP does not provide a way to discard this extra element. In contrast, MP successfully recovers the four GT features---as well as an extra one---for nine runs, and only misses one pixel of one feature for the other run. This is why MP reaches the highest features IOU in Table \ref{table:BD}. The noisy-OR BN trained with MP also learns a prior probability for each feature: the additional feature is always the one with the lowest prior, and can be easily discarded.

\subsection{Reconstructed test images}\label{appendix:BD_rec}
\begin{figure*}[!htbp]
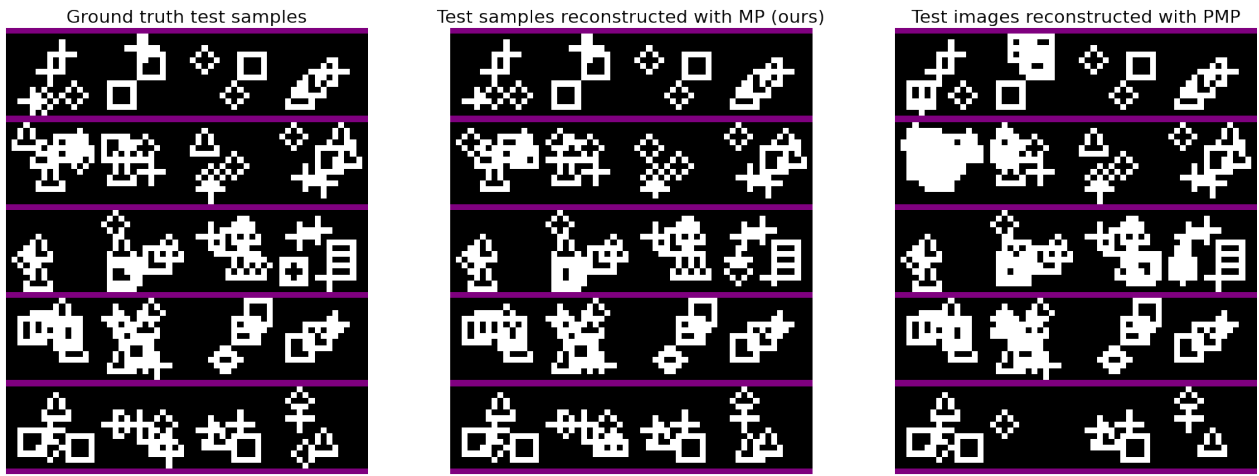

     \centering
     \includegraphics[height=0.38\textwidth]{MP_random_X_test.pdf}
     \hfill
     \includegraphics[height=0.38\textwidth]{MP_random_rec_X_test.pdf}
     \hfill
     \includegraphics[height=0.38\textwidth]{PMP_random_rec_X_test.pdf}
     \caption{
     [Left] Ground truth test scenes for a random seed. [Middle] Test reconstructions returned by our MP method. [Right] Test reconstructions returned by PMP.}
    \label{fig:appendix_BD2}
\end{figure*}

Our last Figure \ref{fig:appendix_BD2} compares the performance of MP and PMP for reconstructing the test scenes on one seed selected at random. We see that PMP performs well, and that our method achieves an almost perfect test reconstruction, which explains that it reaches the lowest test RE in Table \ref{table:BD}, Section \ref{sec:BD}.
